\pdfoutput=1
\documentclass[11pt, table]{article}

\usepackage[preprint]{acl}

\usepackage{times}
\usepackage{latexsym}
\usepackage{enumitem}

\usepackage[T1]{fontenc}
\usepackage{todonotes}
\usepackage[utf8]{inputenc}

\usepackage{microtype}

\usepackage{inconsolata}

\usepackage{bm}
\usepackage{amsmath}

\usepackage{graphicx}
\usepackage{url}
\usepackage{booktabs}

\usepackage{mathtools}
\usepackage{breqn}

\newcommand{\mc}[3]{\multicolumn{#1}{#2}{#3}}

\def\lmatwos{LLaMA-2-7B-Chat}
\def\lmathro{LLaMA-3.2-1B-Instruct}
\def\lmathre{LLaMA-3-8B-Instruct}
\def\gem{Gemma-2B-IT}
\def\gemtwo{Gemma-2-9B-IT}
\def\mis{Mistral-7B-Instruct-v0.2}

\title{DCRM: A Heuristic to Measure Response Pair Quality in Preference Optimization}

\author{
 Chengyu Huang \\
 Cornell University \\
 \texttt{ch2263@cornell.edu} \\\And
 Tanya Goyal \\
 Cornell University \\
 \texttt{tg436@cornell.edu} \\}
\begin{document}
\maketitle
\begin{abstract}

Recent research has attempted to associate preference optimization (PO) performance with the underlying preference datasets. In this work, our observation is that the differences between the preferred response $y^+$ and dispreferred response $y^-$ influence what LLMs \textit{can learn}, which may not match the desirable differences \textit{to learn}. Therefore, we use distance and reward margin to quantify these differences, and combine them to get Distance Calibrated Reward Margin (\texttt{DCRM}), a metric that measures the quality of a response pair for PO. Intuitively, \texttt{DCRM} encourages minimal noisy differences and maximal desired differences. With this, we study 3 types of commonly used preference datasets, classified along two axes: the source of the responses and the preference labeling function. We establish a general correlation between higher \texttt{DCRM} of the training set and better learning outcome. Inspired by this, we propose a \textit{best-of-$N^2$} pairing method that selects response pairs with the highest \texttt{DCRM}. Empirically, in various settings, our method produces training datasets that can further improve models' performance on AlpacaEval, MT-Bench, and Arena-Hard over the existing training sets.\footnote{Our code is at \url{https://github.com/HCY123902/DCRM}.}

\end{abstract}

\section{Introduction}
\label{sec:introduction}

\begin{figure}[t]
    \centering
    \includegraphics[width=0.8\linewidth]{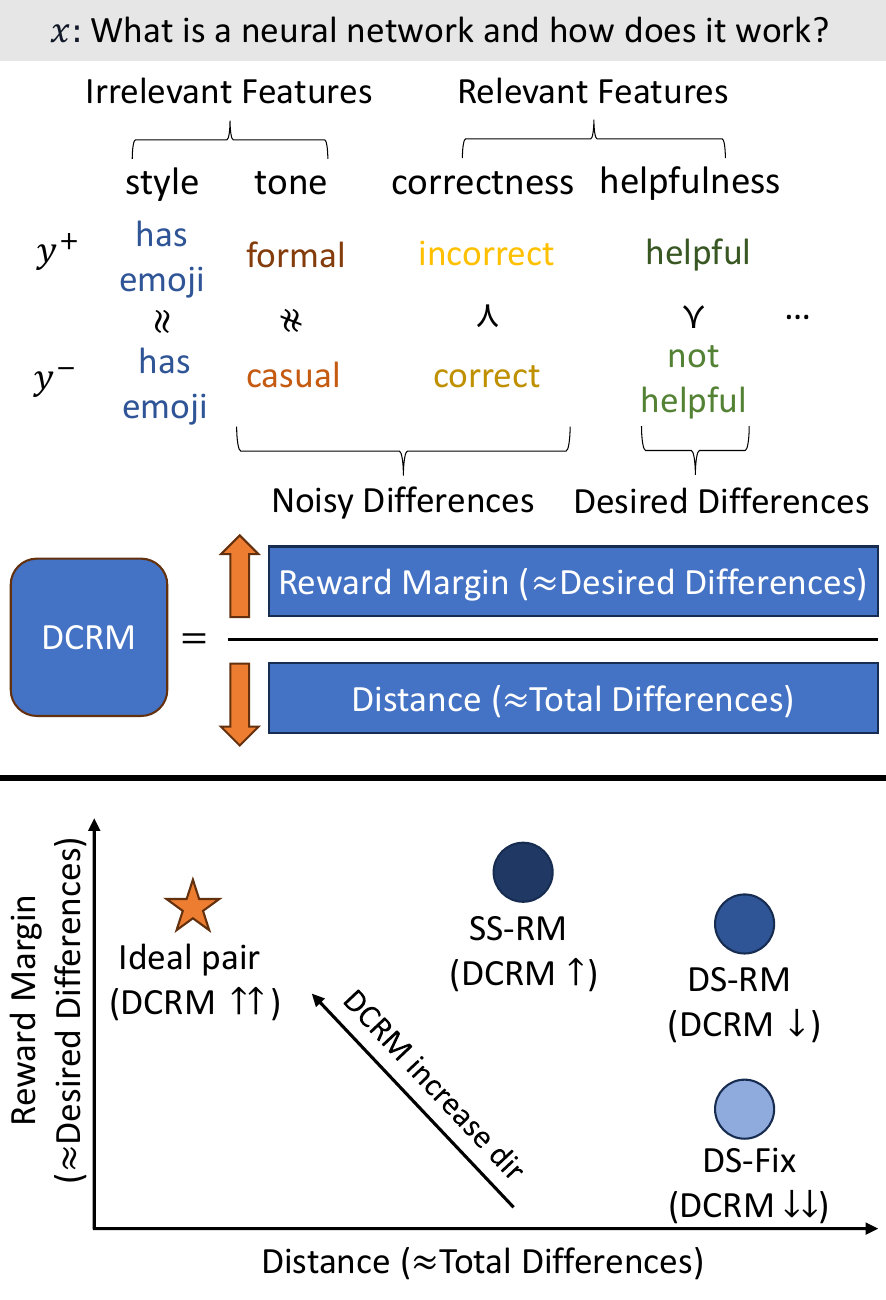}
    \caption{Top: Ideal response pairs should have fewer noisy differences (small distances) and more desired differences (large reward margins). \texttt{DCRM} measures response pair quality with this intuition; Bottom: Common preference datasets (SS-RM, DS-RM, DS-Fix; See \S~\ref{task_setup:dataset}) have varying locations in the distance-reward margin landscape, but none achieves an ideal combination.
    % \tg{think about what you want to say with your fig 1. this is completely irrelevant to DCRM. i would much rather have a catoonish figure of x-axis distance, y-axis - reward and some representation of where we want to sample from and where the different settings we explore end of sampling from.}
    }
    \label{fig:introduction}
\vspace{-1\baselineskip}
\end{figure}

% \begin{figure}[t]
%     \centering
%     \includegraphics[width=0.8\linewidth]{images/Introduction.pdf}
%     \caption{High quality response pairs should have fewer noisy differences and higher proportion of desired differences. Inspired by this intuition, we design \texttt{DCRM}, which measures the density of desired differences. It increases with amount of desired differences and decreases with the amount of total differences. \tg{This is too similar to the APO paper fig. It seemed extremely risky given that we don't even compare to it directly.}}
%     \label{fig:introduction}
% \end{figure}

Preference optimization (PO) methods such as DPO \citep{rafailov2024} have shown success in improving LLMs' performance in various tasks \citep{dubois2024}. These methods usually involve a contrastive learning objective that encourages LLMs to generate a preferred response $y^+$ with higher probability and a dispreferred response $y^-$ with lower probability, given a query $x$. 

Prior research \citep{tang2024,razin2024unintentional} has shown the importance of selecting suitable response pairs for PO training. In particular, the contrastive training signals sent to LLMs are partly derived from the differences between $y^+$ and $y^-$. These differences influence what LLMs \textit{can learn}, which often do not exactly match the set of desirable differences \textit{to learn}. This is because, aside from differences that we want models to learn (useful signals; e.g., $y^+$ is more helpful than $y^-$ in factoid question answering), there can be noisy differences (noisy signals). For instance, $y^+$ and $y^-$ can differ in features that are irrelevant for a task (e.g., different writing styles for factoid question answering) or that the differences are in an incorrect direction (e.g., $y^+$ is less correct than $y^-$). Intuitively, if there are more noisy differences, then LLMs may not effectively learn the desired differences (e.g., to be more helpful) (See Figure~\ref{fig:introduction}).

Although prior research \citep{doosterlinck2024, wu2024} has investigated the correlation between certain proxies of "differences" (e.g., edit distance) and PO learning outcome, it does not distinguish noisy and desired differences, and therefore cannot accurately model the relationship. 
% \tg{we need to engage with APO here. it already has enough enalayis w/ edit distance. we can't not meaningfully engage with it. what is it that we are doing that offers a different insight?}

Therefore, we develop a metric called Distance Calibrated Reward Margin (\texttt{DCRM}) that aims to measure the density of desired differences among the total differences present. \texttt{DCRM} is the ratio between the reward margin, which is a proxy for the amount of desired differences, and two distance metrics (edit distance, probability difference), which are proxies for the total amount of differences.

To study \texttt{DCRM}, we study three common types of preference datasets, categorized by their (1) response sources and (2) preference labeling scheme. We use Ultrafeedback \citep{cui2023} as the seed to construct the datasets, and find that different types of datasets vary in their average \texttt{DCRM} values.

We train three base models (\lmatwos, \lmathro, \gem) on these datasets and use AlpacaEval \citep{dubois2024}, MT-Bench \citep{zheng2023}, and Arena-Hard \citep{li2024} for evaluation. Across all settings, we notice a correlation between higher \texttt{DCRM} and better training outcomes. We further conduct a feature analysis to inspect the properties of each dataset and understand qualitatively what signals (i.e., noisy or desired differences) models learn after training. Inspired by the aforementioned correlation, we propose a method called Best of $N^2$ pairing to select response pairs with high \texttt{DCRM}, and show that training LLMs on the new datasets gives higher performance than on the original datasets. Our contribution is summarized as follows.
% \vspace{-1mm}
\begin{itemize}
    \item We propose a novel metric \texttt{DCRM} that measures the quality of a response pair for PO training.
    % \vspace{-1mm}
    \item We compare three common types of preference datasets and show a positive correlation between the average \texttt{DCRM} value of a training dataset and the training effectiveness.
    % \vspace{-1mm}
    \item We propose \textit{best-of-$N^2$} pairing, which selects response pairs with high \texttt{DCRM} values for effective PO training.
\end{itemize}

\section{Task Setup}
\label{sec:task_setup}

%In the following sections, we first define our task setting in \S~\ref{task_setup:definition}, then cover the general preference datasets that we study in \S~\ref{task_setup:dataset}, and finally introduce the heuristics we use to design DCRM in \S~\ref{task_setup:heuristics}.

\subsection{Problem Definition}
\label{task_setup:definition}

Let $\pi(y|x)$ be a language model (LM) that places a probability distribution over response $y$ conditioned on input $x$. Let $\mathcal{D} = \{x_i, y_i^+, y_i^- \}$ be a preference dataset where responses $y^+$ are preferred to $y^-$. Offline preference optimization, like Direct Preference Optimization (DPO)\footnote{Many variations of DPO have been proposed \citep{azar2023, park2024, meng2024, hong2024}. Since our focus in this work is investigating the impact of preference dataset choices, we fix DPO as our PO algorithm.} \cite{rafailov2024}, use $\mathcal{D}$ to train model $\pi_{\theta}$ starting from the base model $\pi_{\mathrm{ref}}$, by minimizing the following loss: 

\begin{dmath*}
    \mathcal{L}_{DPO} = -E_{(x,y^+, y-)\sim D}\left[\log\sigma \left(\beta \log\frac{\pi_\theta(y^+|x)}{\pi_{\mathrm{ref}}(y^+|x)}-\beta \log\frac{\pi_\theta(y^-|x)}{\pi_{\mathrm{ref}}(y^-|x)}\right)\right]
\end{dmath*}

\noindent where $\beta$ is a hyperparameter.

In this work, we aim to understand how qualitative and quantitative differences between $y^+$ and $y^-$ influence the learning behavior of DPO.

\subsection{Preference Datasets}
\label{task_setup:dataset}

\begin{figure}[h]
    \centering
    \includegraphics[width=\linewidth]{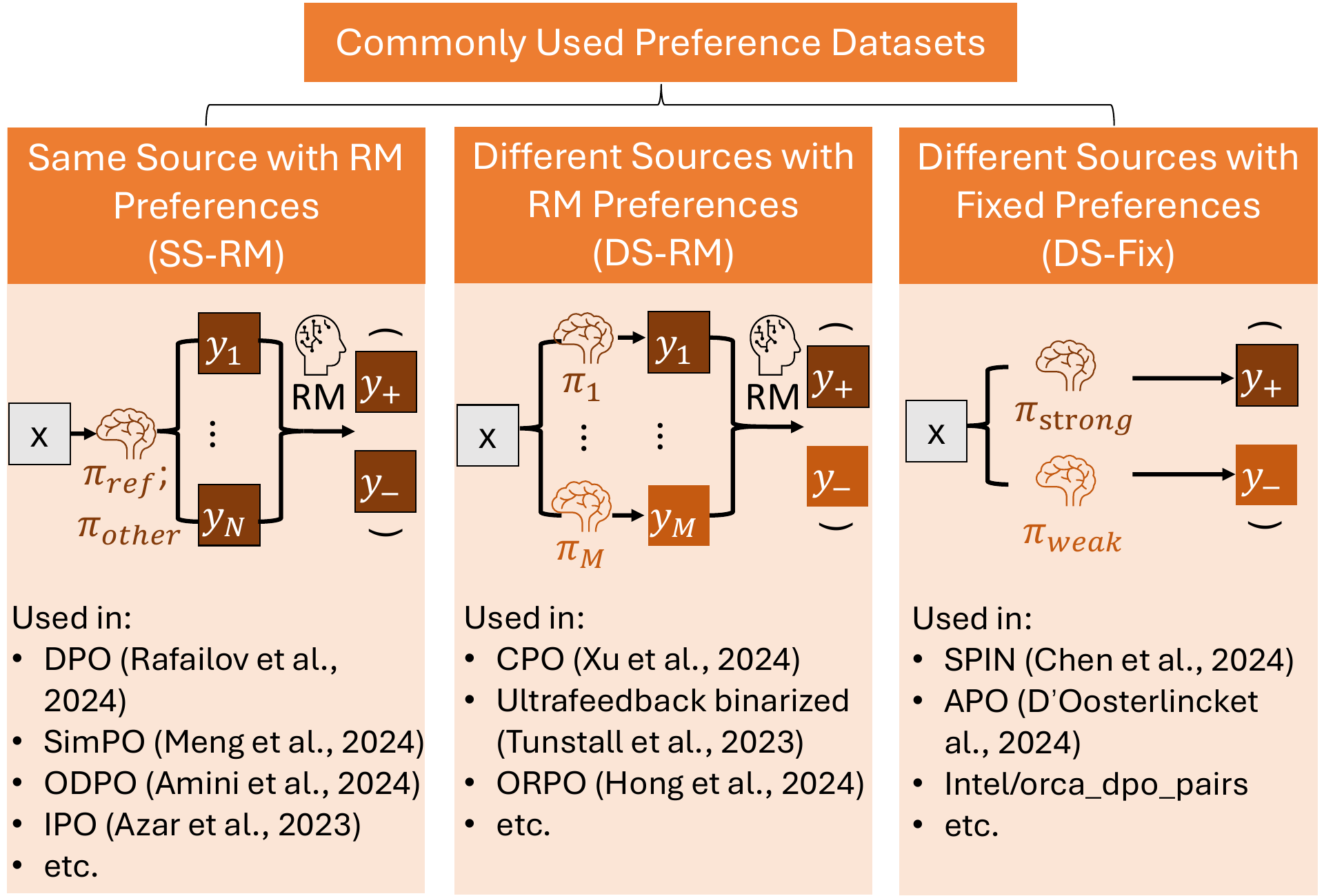}
    \caption{Commonly used preference datasets, categorized into 3 types according to their responses sources and preference labeling functions.}
    \label{fig:preliminary}
\end{figure}

To guide our investigation, we group common techniques for preference dataset curation into 3 categories, according to two axes: source distribution of the response $y$, and the preference labeling function (see Figure~\ref{fig:preliminary}).

\paragraph{\texttt{Same Source w/ RM Preference (SS-RM)}} The original DPO work \citep{rafailov2024} proposed to sample $y^+$ and $y^-$ from the same model, $\pi_{\mathrm{ref}}$ (SS$_{\pi_{\mathrm{ref}}}$), and derive the preference labels using a reward model. %$y^+$ and $y^-$ are set to the response with the highest and lowest reward respectively. 
This has been widely adopted in follow-up works \citep{meng2024, amini2024, azar2023, lai2024}.
% , particularly with the recent performance improvement of learned reward model \citep{wang2024}. 
Note that $y^+$ and $y^-$ can also be from the same source that is not $\pi_{\mathrm{ref}}$ (SS$_{\pi_{\mathrm{other}}}$), meaning that these datasets can be re-used to train a different base LLM too.

\paragraph{\texttt{Diff Source w/ RM Preference (DS-RM)}} Earlier work in DPO used output pairs sampled from two different humans \citep{kopf2023} or models (Ultrafeedback binarized \citep{cui2023, tunstall2023}; Argilla-OpenOrca\footnote{\url{https://huggingface.co/datasets/argilla/distilabel-intel-orca-dpo-pairs}}) to construct the dataset (i.e., $y^+$ is from a different source than $y^-$). The preference labels were typically assigned using a reward model or LLM-based judges. This dataset construction is agnostic to the choice of the policy $\pi_{\mathrm{ref}}$. Once created, these datasets can again be re-used without additional sampling or preference labeling overhead for any new choice of $\pi_{\mathrm{ref}}$ \citep{wu2024, hong2024, bai2022}.

\paragraph{\texttt{Diff Source w/ Fixed Preference (DS-Fix)}} It is possible to have a prior estimate of the relative strengths of two sampling sources (e.g. using rankings on benchmarks like Chatbot-Arena \citep{chiang2024}). In such scenarios, instance-level preference between 2 responses from different sources can be assigned based on model-level rankings (i.e., $y^+$ is always from a "stronger" model than $y^-$). Methods such as SPIN \citep{chen2024} have successfully used such strategies (setting $y^- \sim \pi_{\mathrm{ref}})$ while others \citep{doosterlinck2024} report suboptimal performance with these datasets. 

\subsection{Measuring density of desired differences}
\label{task_setup:heuristics}

Our goal is to study how corpus-level differences in preference pairs impact models' learned behavior after DPO. We quantify the difference between $y^+$ and $y^-$ using a combination of three metrics, which we explain and motivate below: 

%,nolistsep
\paragraph{Token-level edit distance} ($e_\Delta$) between $y^+$ and $y^-$ is the first distance metric that we use. It is the token-level Levenshtein distance between 2 outputs. $e_\Delta$ is easily computable and $\pi_{\mathrm{ref}}$ agnostic. It captures differences in length, lexicon, syntax, etc. 

\paragraph{$\bm{\pi_{\mathrm{ref}}}$'s LogProb Difference} ($p_\Delta$) is the second distance metric that we use. It is computed as $|\log\pi_{\mathrm{ref}}(y^+ | x) - \log\pi_{\mathrm{ref}}(y^- | x)|$. $p_\Delta$ measures the difference in probability mass placed on $y^+$ and $y^-$ by $\pi_{\mathrm{ref}}$. It captures a different notion of ``distance'' from edit-distance; two samples can be very different lexically but be assigned similar probability by $\pi_{\mathrm{ref}}$, or vice versa. These are tougher for the implicit reward model in DPO to distinguish, and this measure helps us account for such instances.

\paragraph{Reward Margin} ($r_\Delta$) measures the difference in rewards from a reward model $\mathrm{RM}$. It is computed as $r_\Delta = r_{y^+} - r_{y^-}$, where $r_y$ is the reward score $\mathrm{RM}$ assigns to an output $y$. This reward margin quantifies the desired differences in targeted (relevant) features between the two outputs, irrespective of their lexical and probability differences.

We combine these to construct a single metric that measures the density of ``desired'' differences between two outputs. We call this \textbf{distance-calibrated reward margin} (\texttt{DCRM}): 
\begin{align}
    \mathrm{DCRM}(y^+, y^-) = \frac{\sigma(r_\Delta) - 0.5}{e_\Delta+p_\Delta +\epsilon}
\end{align}
We omit $(y^+, y^-)$ as the arguments for $r_\Delta,e_\Delta, p_\Delta$ for brevity and include constant $\epsilon =1$ for numeric stability. The numerator captures the normalized reward margin\footnote{We apply the sigmoid function to normalize $r_\Delta$ to be between [0, 1] and subtract 0.5 to preserve the margin sign.} between $y^+$ and $y^-$ (a 0-centered Bradley-Terry model \citep{bradley1952}), and the denominator measures their distances (i.e., lexical and probabilistic differences).\footnote{We do not adjust the scales of $e_\Delta$ and $p_\Delta$ since we find that these are similar across most settings in our experiments.}

We hypothesize that when the useful contrast signals (desired differences, measured by $r_\Delta$) are a large fraction of the total differences (measured by $e_\Delta + p_\Delta$) in the response pair (i.e., useful signals are dense), training becomes more effective.

\texttt{DCRM} captures this hypothesis. A high \texttt{DCRM} implies (1) a high reward margin between $y^+$ and $y^-$ (i.e. there are many desired differences between the two for $\pi_{\mathrm{ref}}$ to learn from) and (2) low distances between the two (i.e., the total differences are small). 
% This could be because of low edit distance (desired differences as localized to short spans) or low log probability difference ($\pi_{\mathrm{ref}}$ does not place a substantially higher probability on $y^+$). 
In this case, training signals are more meaningful and less noisy for the LLMs to learn effectively.\footnote{See Appendix~\ref{app:dcrm_properties} for the properties of \texttt{DCRM}.}

\section{Experiment Setup}
\label{sec:experiment_setup}

\subsection{Training Setup}
\paragraph{Models} We experiment with three options for our base model ($\pi_{\mathrm{ref}}$). They include LLaMA2 (\lmatwos; \citet{touvron2023}), LLaMA3.2 (\lmathro; \citet{grattafiori2024}, and an extra model from other series Gemma (\gem; \citet{mesnard2024}). We train each of these models using the DPO objective for 2 epochs, and select the best checkpoint based on validation performance. Please refer to Appendix~\ref{app:training_details} for other training details. Due to length constraints, we report results for LLaMA2 and LLaMA3.2 in the main paper, and put the results for Gemma in Appendix~\ref{app:complete_results}.

We use the overall scores from the reward model ArmoRM \cite{wang2024} to compute $r\Delta$.

\paragraph{Preference Datasets}
\label{sec:dataset}

% \begin{table*}[t]
%   \centering
%   \small
%   %\resizebox{\textwidth}{!}{
%   \begin{tabular}{llcccc}
%     \toprule
%     Type & Dataset & $e_\Delta$ & $p_\Delta$ &$r_\Delta$(e-2)& DCRM(e-2)\\
%     \midrule
%     SS-RM & $\pi_{\mathrm{ref}}=$LLaMA2 & 427 & \textbf{32.48} & 2.815 & 4.542 \\
%          & $\pi_{\mathrm{ref}}$=LLaMA3.2 & 434 & 120.07 & \textbf{4.220} & \textbf{7.534} \\
%        & Gma2 & \textbf{370} & 91.78/84.78 & 1.703 & 2.865/3.154 \\
%        & Mst & 526 & 158.54/176.22 & 2.129 & 1.589/1.680 \\
%     \midrule
%     DS-RM & Gma2-Mst & 542 & 226.47/228.22 & 2.026 & 1.127/1.171 \\
%     \midrule
%     DS-Fix & Gma2-Mst & 542 & 226.47/228.22 & 1.016 & 0.429/0.441\\
%     \bottomrule
%   \end{tabular}
%   \caption{\label{tab:dataset_statistics}
%     Summary of the datasets. Numbers left of / are measured with llama-2-7b-chat and right of / are measured with llama-3.2-1b-inst. The reported DCRM values are scaled 1k times for visualization, which does not affect correlation analysis. SS-RM datasets have the highest DCRM while DS-Fix ones have the lowest DCRM. 
%     % \tg{cant have two different naming conventions for gemma2 in the same table (G2 vs Gemma).} \tg{I posted some "good practices" in the lab slack. All numbers in a table ideally, but defniitely in the same column shuold hve the same number of decimal places. fix for the last column.} \tg{We don't need the y+ and y- columns. these add redundant info. the only extra info i got was that for DS-RM, 70\% of the poistive samples are from gemma2 which can easily be a part of the text.}
%   }
% \end{table*} 

\begin{table}[t]
  \centering
  \small
  %\resizebox{\textwidth}{!}{
  \begin{tabular}{llcccc}
    \toprule
    Type & Dataset & $e_\Delta$ & $p_\Delta$ &$r_\Delta${\tiny (e-2)}& \texttt{DCRM}{\tiny (e-2)}\\
    \rowcolor{lightgray}\mc{6}{c}{$\pi_{\mathrm{ref}}=$LLaMA2 (\textit{\lmatwos})}\\
    SS-RM & $\pi_{\mathrm{ref}}$ & 427 & \textbf{32.48} & \textbf{2.82} & \textbf{4.54} \\
       & Gma2 & \textbf{370} & 91.78 & 1.70 & 2.87 \\
       & Mst & 526 & 158.54 & 2.13 & 1.59 \\
    \midrule
    DS-RM & Gma2-Mst & 542 & 226.47 & 2.03 & 1.13 \\
    \midrule
    DS-Fix & Gma2-Mst & 542 & 226.47 & 1.02 & 0.43\\
    \rowcolor{lightgray}\mc{6}{c}{$\pi_{\mathrm{ref}}=$LLaMA3.2 (\textit{\lmathro})}\\
    SS-RM & $\pi_{\mathrm{ref}}$ & 434 & 120.07 & \textbf{4.22} & \textbf{7.53} \\
       & Gma2 & \textbf{370} & \textbf{84.78} & 1.70 & 3.15 \\
       & Mst & 526 & 176.22 & 2.13 & 1.68 \\
    \midrule
    DS-RM & Gma2-Mst & 542 & 228.22 & 2.03 & 1.17 \\
    \midrule
    DS-Fix & Gma2-Mst & 542 & 228.22 & 1.02 & 0.44\\
    \bottomrule
  \end{tabular}
  \caption{\label{tab:dataset_statistics}
    Statistics of the datasets. Each metric value is averaged across examples. Changing $\pi_{\mathrm{ref}}$ changes $p_\Delta$ and so we report separate statistics for LLaMA2 and LLaMA3.2. The reported \texttt{DCRM} values are scaled 1k times for visualization, which does not affect correlation analysis. SS-RM datasets have the highest \texttt{DCRM} while DS-Fix ones have the lowest \texttt{DCRM}.
  }
  \vspace{-\baselineskip}
\end{table}

We use the 60K prompts from Ultrafeedback \citep{cui2023}. We create our preference datasets using responses sampled from four different models across the three settings (\texttt{SS-RM}, \texttt{DS-RM}, \texttt{DS-Fix}) described in \S~\ref{task_setup:dataset}. 

For \textbf{\texttt{SS-RM}}, we sample responses from the base model $\pi_{\mathrm{ref}}$. We also use \gemtwo~(Gma2) and \mis~(Mst) as two extra sources of responses. For each source, we follow \citet{meng2024} and sample $N=5$ responses and then select the best response pair with the highest $r_\Delta$ using the reward model $\mathrm{RM}$.

For \textbf{\texttt{DS-RM}}, we fix the source distributions to \gemtwo~(Gma2) and \mis~(Mst). We sample one response from each, and decide the preference label using $\mathrm{RM}$. We find that roughly 70\% of $y^+$ comes from Gma2 and 70\% of $y^-$ comes from Mst.

For \textbf{\texttt{DS-Fix}}, we use the same response pairs as \texttt{DS-RM}, but always set $y^+$ to be from \gemtwo~(stronger model) and $y^-$ to be from \mis~(weaker model), respectively.

\paragraph{Dataset Statistics}
Table~\ref{tab:dataset_statistics} shows the dataset statistics. As expected, \texttt{SS-RM} datasets, which get the paired responses from the same source, have the lowest $e_\Delta$ and $p_\Delta$, leading to the highest overall \texttt{DCRM}. \texttt{DS-RM} has higher distances and consequently lower \texttt{DCRM}. Surprisingly, we find that \texttt{DS-Fix} has the lowest reward margin even though its samples have a higher lexical difference. This makes it have the lowest \texttt{DCRM} across the three settings.

% \tg{DS-RM also has \textit{large distances} but additionally its reward margin is smaller due to the lack of a reward model. <-- previous wording i changed, i get this but the reader wont. the reason is that becuase there isnt a reward model, the + and the - keep adding up to near zero diff, but the wording makes it seem like reward model pushed samples apart. }

\subsection{Quantitative Evaluation}
\label{sec:eval_metrics}

We evaluate the general conversational and instruction-following abilities of our trained models $\pi_\theta$ using three chat benchmarks, AlpacaEval, MT-Bench, and Arena-Hard. AlpacaEval reports the models' win rates against a baseline model, GPT-4-1106-Preview \citep{achiam2024}. Arena-Hard run similar evaluations, with GPT-4-0314 as the baseline model. MT-Bench is a multi-turn conversational benchmark and uses a judge model to score the model's responses on a scale of 10.\footnote{For all three benchmarks, we use GPT-4o-mini-2024-0718 \cite{hurst2024} as the judge to regulate costs.}

% We note that this GPT-4 model is substantially stronger than our models. Comparisons against it may not surface differences between similar capability models. Therefore, we also report results using the same setup but with $\pi_{\mathrm{ref}}$ as the baseline instead.
%For more task-specific and out-of-distribution downstream performance, we evaluate our model on GSM8K \citep{cobbe2021} and MixEval-Hard \cite{ni2024} and report the results in Appendix~\ref{app:downstream_performance}.

\section{Comparing Different Types of Preference Datasets}
\label{sec:result}
In this section, we compare models that are trained on different types of preference datasets, and establish a correlation between the dataset-level \texttt{DCRM} value and downstream performances. We report the results in Table~\ref{tab:main_res}.

\begin{table}
  \small
  \centering
  \begin{tabular}{llcc|c|c}
    \toprule
     & & AP-L & AP-R & MT & AH \\
    \midrule
     & LLaMA2 & 12.57 & 10.43 & 5.41 & 8.90 \\
    \midrule
    SS-RM & +$\pi_{\mathrm{ref}}$ & \textbf{22.36} & \textbf{16.81} & 5.55 & \textbf{16.67} \\
     & +Gma2      & 15.89 & 13.12 & \textbf{5.50} & 11.57 \\
     & +Mst     &  15.49 & 12.07 & 5.40 & 10.42 \\
    \midrule
    DS-RM & +Gma2-Mst & 14.13 & 11.51 & 5.52 & 10.55 \\
    \midrule
    DS-Fix & +Gma2-Mst & 13.26 & 8.99 & 5.24 & 6.68 \\
    \midrule \midrule
     & LLaMA3.2 & 14.15 & 15.34 & 4.66 & 10.88 \\
    \midrule
    SS-RM & +$\pi_{\mathrm{ref}}$ & 22.80 & 25.65 & \textbf{5.01} & \textbf{18.88} \\
     & +Gma2      & \textbf{24.57} & \textbf{27.52} & 4.99 & 15.91 \\
     & +Mst     & 19.43 & 19.94 & 4.91 & 16.03 \\
    \midrule
    DS-RM & +Gma2-Mst & 20.01 & 21.61 & \textbf{5.01} & 13.61 \\
    \midrule
    DS-Fix & +Gma2-Mst & 10.31 & 8.20 & 4.54 & 14.94 \\
    \bottomrule
  \end{tabular}
  \caption{\label{tab:main_res}
    Main Results; AP-L: Length-Controlled Win Rate on AlpacaEval; AP-R: Raw Win Rate on AlpacaEval; MT: MT-Bench Score; AH: Arena-Hard Win Rate;
    % LC$_{\mathrm{ref}}$ and WR$_{\mathrm{ref}}$: Win rates against $\pi_{\mathrm{ref}}$ instead of GPT-4-1106-Preview.
    SS-RM datasets generally lead to the best performance while DS-Fix ones lead to the worst performance. 
    % \tg{can we have this as a two column table instead, since it is overflowing right now?}\ch{I added a type column, if that is what you asked for.}
  }
  \vspace{-\baselineskip}
\end{table}

% \begin{table}
%   \small
%   \centering
%   \begin{tabular}{llcc|cc}
%     \toprule
%      & & LC & WR & LC$_{\mathrm{ref}}$ & WR$_{\mathrm{ref}}$ \\
%     \midrule
%      & LLaMA2 & 12.57 & 10.43 & 50 & 50 \\
%     \midrule
%     SS-RM & +$\pi_{\mathrm{ref}}$ & \textbf{22.36} & \textbf{16.81} & \textbf{71.66} & \textbf{70.06} \\
%      & +Gma2      & 15.89 & 13.12 & 58.40 & 58.95 \\
%      & +Mst     &  15.49 & 12.07 & 60.86 & 58.22 \\
%     \midrule
%     DS-RM & +Gma2-Mst & 14.13 & 11.51 & 56.29 & 56.15 \\
%     \midrule
%     DS-Fix & +Gma2-Mst & 13.26 & 8.99 & 49.72 & 42.67 \\
%     \midrule \midrule
%      & LLaMA3.2 & 14.15 & 15.34 & 50 & 50 \\
%     \midrule
%     SS-RM & +$\pi_{\mathrm{ref}}$ & 22.80 & 25.65 & \textbf{67.89} & \textbf{70.00} \\
%      & +Gma2      & \textbf{24.57} & \textbf{27.52} & 67.09 & 69.44 \\
%      & +Mst     & 19.43 & 19.94 & 61.29 & 60.06 \\
%     \midrule
%     DS-RM & +Gma2-Mst & 20.01 & 21.61 & 56.92 & 57.39 \\
%     \midrule
%     DS-Fix & +Gma2-Mst & 10.31 & 8.20 & 34.37 & 26.71 \\
%     \bottomrule
%   \end{tabular}
%   \caption{\label{tab:main_res}
%     Results on AlpacaEval; LC: Length-Controlled Win Rate; WR: Raw Win Rate; LC$_{\mathrm{ref}}$ and WR$_{\mathrm{ref}}$: Win rates against $\pi_{\mathrm{ref}}$ instead of GPT-4-1106-Preview. SS-RM datasets generally lead to the best performance while DS-Fix ones lead to the worst performance. 
%     % \tg{can we have this as a two column table instead, since it is overflowing right now?}\ch{I added a type column, if that is what you asked for.}
%   }
% \end{table}

\paragraph{Sampling from the same source distribution (\texttt{SS-RM}) outperforms other methods.}
Table~\ref{tab:main_res} shows that sampling response pairs from the same distribution ($\pi_{\mathrm{ref}}$ and others) and deriving preferences using the reward model perform better than \texttt{DS-RM} and \texttt{DS-Fix}. In particular, training with responses from ${\pi_{\mathrm{ref}}}$ gives the best performance, which mirrors findings from prior work \citep{tang2024}. Relating back to Table~\ref{tab:dataset_statistics}, \texttt{SS-RM} datasets also have the highest \texttt{DCRM} value. 
% that recommends sampling pairs from $\pi_{\mathrm{ref}}$.

To our surprise, \texttt{SS-RM} Gma2 is on par with \texttt{SS-RM} $\pi_{\mathrm{ref}}$ when $\pi_{\mathrm{ref}}$=LLaMA3.2. Consulting Table~\ref{tab:dataset_statistics}, we see that \texttt{SS-RM} Gma2 has a lower $p_\Delta$ than that of LLaMA3.2, possibly explaining this result.

\paragraph{\texttt{DS-Fix} performs worse than the base model.} This technique
% , although used successfully in prior work \citep{chen2024}, 
performs the worst among the three dataset settings. Similar results have also been reported by \citet{doosterlinck2024}. In fact, we find that its performance is worse than even the starting model. In Appendix~\ref{app:preliminary}, we show that there are consistent stylistic differences between the two source distributions (e.g. presence of more emojis in $Y^+$ than $Y^-$), which is reflected in the model's output after training. Again, relating back, \texttt{DS-Fix} datasets also have the lowest \texttt{DCRM} value.

\begin{figure}
    \centering
    \includegraphics[width=0.9\linewidth]{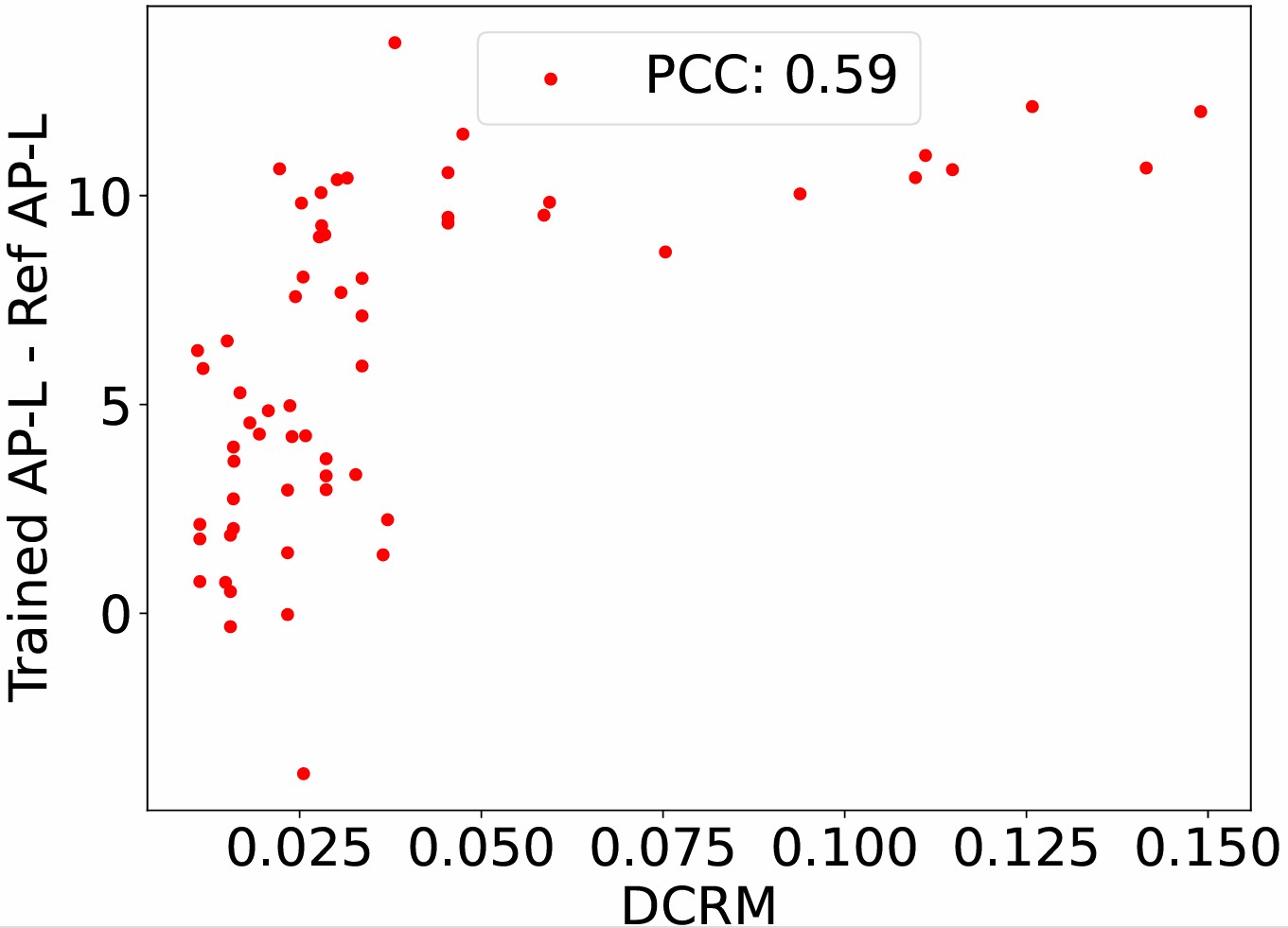}
    \caption{\texttt{DCRM} is positively correlated with models' performance boost on AP-L. PCC: Pearson Correlation Coefficient; Y axis: change in AP-L after training. Each point in the diagram corresponds to a trained model.
    }
    \label{fig:corr_dcrm}
    \vspace{-\baselineskip}
\end{figure}

\paragraph{DCRM is positively correlated with model performance after training.} With the above observations, we formally quantify the correlation between \texttt{DCRM} and downstream performance. To include sufficient data points, we sample multiple outputs from the source distributions and select response pairs that vary the dataset-level $p_\Delta$, $e_\Delta$, and $r_\Delta$.\footnote{See Appendix~\ref{app:correlation_analysis} for details.} We compute the performance boost, i.e. the AP-L improvement of $\pi_\theta$ against $\pi_{\mathrm{ref}}$, and show its correlation with \texttt{DCRM} in Figure~\ref{fig:corr_dcrm}.\footnote{See MT and AH correlations in Appendix~\ref{app:corr_mt_ah}.}

We find that \texttt{DCRM} and downstream performance are moderately positively correlated, with a Pearson Correlation of $0.59$, which is stronger than the individual metrics -- correlation with $e_\Delta$, $p_\Delta$, and $r_\Delta$ is -0.51, -0.55, and 0.43 respectively (See Appendix~\ref{app:corr_individual_metrics}). We observe a saturation effect once \texttt{DCRM} passes 0.075, and suspect this to be caused by the inherent limitations of the reward model.

\section{Operationalizing DCRM}
\label{sec:implication}

In \S~\ref{sec:result}, we observe that higher \texttt{DCRM} is correlated with better training outcomes. Can we use this correlation to guide training dataset selection?

% \tg{i would usually avoid statemetns like "an obvous solution" "a natural solution" etc. i removed some where i saw. }

\paragraph{Approach}
An answer is to sample responses from $\pi_{\mathrm{ref}}$. However, this can be expensive with a large model or dataset. Instead, we want to investigate how to \textit{select} the best response pair from an \textit{existing} pool of responses,
% so that we can achieve better performance without incurring additional sampling costs.
Formally, given $N$ responses $\{y_1, \cdots, y_N\}$ (and also $\{y_{N+1}, \cdots, y_{2N}\}$ from a second model in the DS setting), we propose to select the pair $(y_i, y_j)$ with the highest \texttt{DCRM}. We denote this as Best of $N^2$ pairing (Bo$N^2$), since we select the best pair from $N\times N$ candidates. Our method is different from the conventional method (used in \texttt{SS-RM}), which chooses the pair with the highest reward margin by setting $y^+$ and $y^-$ to the response with the highest and lowest reward scores.

\paragraph{Setup}
We apply our method to three baselines. In the Same Source (\texttt{SS-RM}) setting, we reselect the response pair using the existing $N$ responses sampled from (1) $\pi_{\mathrm{ref}}$, or (2) Mst. In the Different Sources (\texttt{DS-RM})\footnote{Applying our method to the DS-Fix setting leads to the same dataset as DS-RM, so we combine them together} setting, we use (3) Gma2-Mst as the third baseline, and select a response pair with the highest \texttt{DCRM} while maintaining the condition that $y^+$ and $y^-$ come from different sources.\footnote{Baseline (3) is not strictly a fair comparison. In Appendix~\ref{app:complete_results} we provide a fair baseline w/ $BoN^2$ ($r_\Delta$ only).}
% \tg{very inconsistent notation throughout. Also, I don't know if we need to say we sample N from one and N from another if this is basically described in the prev paragraph.}\ch{Changed}

\begin{table}[t]
  \small
  \centering

  \begin{tabular}{llcccc}
    \toprule
    Type & Dataset & $e_\Delta$ & $p_\Delta$ &$r_\Delta${\tiny (e-2)}& \texttt{DCRM}{\tiny (e-2)}\\
    \rowcolor{lightgray}\mc{6}{c}{$\pi_{\mathrm{ref}}=$LLaMA2 (\textit{\lmatwos})}\\
    SS-RM & $\pi_{\mathrm{ref}}$ & 427 & 32.48 & \textbf{2.82} & 4.54 \\
    & w/ $BoN^2$ & \textbf{370} & \textbf{23.87} & 2.52 & \textbf{5.94} \\
    \midrule
    SS-RM & Mst & 526 & 158.54 & \textbf{2.13} & 1.59 \\
    & w/ $BoN^2$ & \textbf{410} & \textbf{79.94} & 1.79 & \textbf{2.07} \\
    \midrule
    DS-RM & Gma2-Mst & 542 & 226.47 & 2.03 & 1.13 \\
    & w/ $BoN^2$ & \textbf{458} & \textbf{142.94} & \textbf{3.27} & \textbf{2.58} \\
    \rowcolor{lightgray}\mc{6}{c}{$\pi_{\mathrm{ref}}=$LLaMA3.2 (\textit{\lmathro})}\\
    SS-RM & $\pi_{\mathrm{ref}}$ & 434 & 120.07 & \textbf{4.22} & 7.53 \\
    & w/ $BoN^2$ & \textbf{356} & \textbf{63.55} & 3.58 & \textbf{11.48} \\
    \midrule
    SS-RM & Mst & 526 & 176.22 & \textbf{2.13} & 1.68 \\
    & w/ $BoN^2$ & \textbf{339} & \textbf{78.81} & 1.78 & \textbf{2.44} \\
    \midrule
    DS-RM & Gma2-Mst & 542 & 228.22 & 2.03 & 1.17 \\
    & w/ $BoN^2$ & \textbf{374} & \textbf{134.94} & \textbf{3.24} & \textbf{3.02} \\
    \bottomrule
  \end{tabular}
  \caption{\label{tab:new_dataset_statistics}
    Statistics of the original and new datasets; w/ $BoN^2$ indicates datasets whose response pairs are reselected using best-of-$N^2$ method. They have a higher \texttt{DCRM} value than their original counterparts.
  }
  \vspace{-\baselineskip}
\end{table}

Table~\ref{tab:new_dataset_statistics} gives a comparison between the original and reselected datasets. After reselection with \texttt{DCRM}, both $e_\Delta$ and $p_\Delta$ decrease, while $r_\Delta$ stays in a reasonable range without too much drop.

%In \S~\ref{sec:new_main_res},  \

\subsection{Main Results}
\label{sec:new_res}
We compare $BoN^2$ against the baselines in Table~\ref{tab:new_res}.

\begin{table}[t]
  \small
  \centering

  \begin{tabular}{llcc|c|c}
    \toprule
    & & AP-L & AP-R & MT & AH \\
    \midrule
    & LLaMA2 & 12.57 & 10.43 & 5.41 & 8.90 \\
    \midrule
    SS-RM & +$\pi_{\mathrm{ref}}$ & 22.36 & 16.81 & 5.55 & \textbf{16.67} \\
    & w/ $BoN^2$ & \textbf{22.41} & \textbf{17.2} & \textbf{5.67} & 16.07 \\
    \midrule
    SS-RM & +Mst     &  15.49 & 12.07 & 5.40 & 10.42 \\
    & w/ $BoN^2$     &  \textbf{17.42} & \textbf{13.29} & \textbf{5.48} & \textbf{10.99} \\
    \midrule
    DS-RM & +Gma2-Mst & 14.13 & 11.51 & \textbf{5.52} & 10.55 \\
    & w/ $BoN^2$ & \textbf{16.82} & \textbf{13.6} & 5.48 & \textbf{11.75} \\
    \midrule
    \midrule
    & LLaMA3.2 & 14.15 & 15.34 & 4.66 & 10.88 \\
    \midrule
    SS-RM & +$\pi_{\mathrm{ref}}$ & 22.80 & 25.65 & 5.01 & 18.88 \\
    & w/ $BoN^2$ & \textbf{24.77} & \textbf{27.64} & \textbf{5.10} & \textbf{20.25} \\
    \midrule
    SS-RM & +Mst     & 19.43 & 19.94 & 4.91 & 16.03 \\
    & w/ $BoN^2$     &  \textbf{21.73} & \textbf{21.37} & \textbf{5.11} & \textbf{16.62} \\
    \midrule
    DS-RM & +Gma2-Mst & 20.01 & 21.61 & 5.01 & 13.61 \\
    & w/ $BoN^2$ & \textbf{24.53} & \textbf{27.76} & \textbf{5.04} & \textbf{19.17} \\
    \bottomrule
  \end{tabular}
  \caption{\label{tab:new_res}
    Main Results; $BoN^2$ datasets give a stronger performance than their original counterparts. 
  }
  \vspace{-1.5\baselineskip}
\end{table}

% \begin{table}[t]
%   \small
%   \centering

%   \begin{tabular}{llcc|cc}
%     \toprule
%     & & LC & WR & LC$_{\mathrm{ref}}$ & WR$_{\mathrm{ref}}$ \\
%     \midrule
%     & LLaMA2 & 12.57 & 10.43 & 50 & 50 \\
%     \midrule
%     SS-RM & +$\pi_{\mathrm{ref}}$ & 22.36 & 16.81 & 71.66 & 70.06 \\
%     & w/ $BoN^2$ & \textbf{22.41} & \textbf{17.2} & \textbf{72.92} & \textbf{72.30} \\
%     \midrule
%     SS-RM & +Mst     &  15.49 & 12.07 & \textbf{60.86} & \textbf{58.22} \\
%     & w/ $BoN^2$     &  \textbf{17.42} & \textbf{13.29} & 59.18 & 57.08 \\
%     \midrule
%     DS-RM & +Gma2-Mst & 14.13 & 11.51 & 56.29 & 56.15 \\
%     & w/ $BoN^2$ & \textbf{16.82} & \textbf{13.6} & \textbf{61.45} & \textbf{60.93} \\
%     \midrule
%     \midrule
%     & LLaMA3.2 & 14.15 & 15.34 & 50 & 50 \\
%     \midrule
%     SS-RM & +$\pi_{\mathrm{ref}}$ & 22.80 & 25.65 & 67.89 & 70.00 \\
%     & w/ $BoN^2$ & \textbf{24.77} & \textbf{27.64} & \textbf{69.11} & \textbf{71.49} \\
%     \midrule
%     SS-RM & +Mst     & 19.43 & 19.94 & 61.29 & \textbf{60.06} \\
%     & w/ $BoN^2$     &  \textbf{21.73} & \textbf{21.37} & \textbf{61.70} & 59.25 \\
%     \midrule
%     DS-RM & +Gma2-Mst & 20.01 & 21.61 & 56.92 & 57.39 \\
%     & w/ $BoN^2$ & \textbf{24.53} & \textbf{27.76} & \textbf{68.40} & \textbf{69.38} \\
%     \bottomrule
%   \end{tabular}
%   \caption{\label{tab:new_main_res}
%     Results on AlpacaEval; $BoN^2$ datasets give a stronger performance than their original counterparts. 
%   }
% \end{table}

\vspace{-1mm}
\paragraph{Best of $N^2$ pairing increases performance across all settings.} When training LLaMA3.2, we observe a higher performance across all baselines. When training LLaMA2, performance increases notably on top of both Mst (SS-RM) and Gma2-Mst (DS-RM), especially for the latter.
% \tg{this was difficult for me to parse since there is a switch between talking about $pi_{ref}$ settings and them about ss-rm. can you take another pass at this which re-write this in the same style as the section above. }\ch{Changed}
\vspace{-1mm}

However, performance only increases marginally in the LLaMA2 $\pi_{\mathrm{ref}}$ (SS-RM) setting. We suspect that most responses from LLaMA2 are similar to each other. In this case, maximizing the reward margin will not incur very high distances, so the response pairs from $\pi_{\mathrm{ref}}$ (SS-RM) are already close to the best. There is little room for improvement no matter how we reselect the pairs. This is evident in Table~\ref{tab:new_dataset_statistics}, where we observe a smaller reduction in $e_\Delta$ and $p_\Delta$ compared with every other setting.

% \tg{the way i would re-write this subsection is this. first paragraph: BoN2 improves performace when samples are from difference distributions. in fact, for ds-rm g2-m we see that we are able to match the performance of the ss-rm pi ref settings for llama3.2 .  second paragraph: we see more marginal improvements in the SS-RM setting. and then talk about llama3.2 vs llama2}

\subsection{Ablation Study}

\begin{table}[h]
  \small
  \centering

  \begin{tabular}{lcc|c|c}
    \toprule
     & AP-L & AP-R & MT & AH \\
    \midrule
    LLaMA2 & 12.57 & 10.43 & 5.41 & 8.90 \\
    \midrule
    +$\pi_{\mathrm{ref}}$ & 22.36 & 16.81 & 5.55 & \textbf{16.67} \\
    w/ $BoN^2$ & 22.41 & 17.20 & \textbf{5.67} & 16.07 \\
    ~~~~-$p_\Delta$  & 22.1 & \textbf{17.27} & 5.59 & 15.62 \\
    ~~~~-$e_\Delta$  & \textbf{24.04} & 17.14 & 5.51 & 14.61 \\
    ~~~~-$r_\Delta$ & 14.81 & 12.11 & 5.54 & 12.97 \\
    \bottomrule
  \end{tabular}
  \caption{\label{tab:ablation}
    Ablation Study on \texttt{DCRM} in the SS-RM setting; Removing $p_\Delta$ or $e_\Delta$ hurts performance slightly, while removing $r_\Delta$ significantly reduces performance. 
    % \tg{if everything is SS-rM why have it here in the table and not just the caption?}
  }
  \vspace{-\baselineskip}
\end{table}

Since \texttt{DCRM} is composed of three metrics, we do an ablation study of our method in the $\pi_{\mathrm{ref}}$ (SS-RM) setting. We remove one of $p_\Delta$, $e_\Delta$, or $r_\Delta$ from \texttt{DCRM} and reselect the response pair. Table~\ref{tab:ablation} shows that \textbf{removing $p_\Delta$ gives a performance close to that of the complete metric, while removing $e_\Delta$ slightly hurts performance}. In Appendix~\ref{app:ablation_study}, we show that removing either of these in the Mst (SS-RM) and DS-RM settings can still give a performance boost over the original datasets, which means in these settings our method can be effective with a cheaper computation.

\paragraph{Removing $r_\Delta$ makes training much less effective.} This is expected, since without $r_\Delta$ our method selects response pairs that have the smallest distances and are minimally different. This not only eliminates noisy differences, but also those useful ones.

\section{Qualitative Analysis (Feature-Analysis)}
\label{sec:qualitative_analysis}
\S~\ref{sec:result} and \S~\ref{sec:implication} show the correlation between the \texttt{DCRM} value of a training set and \textit{quantitative} performance. We also want to inspect whether these datasets have \textit{qualitative} differences, to validate our starting motivation that connects performance with data quality (i.e., more desired differences and fewer noisy ones between $y^+$ and $y^-$ make PO more effective), and better ground \texttt{DCRM} with this quality.

We analyze the feature differences between $y^+$ and $y^-$. We define \textit{relevant} features (correctness, helpfulness, etc.) as those that the LLMs should learn, and \textit{irrelevant} features (writing style, sarcasm, tone, etc.) as those not targeted by the task.

\paragraph{Features} To align with the reward signals, we use the 11 features (de-duplicated) from the ArmoRM reward model as the relevant features. These include helpfulness, truthfulness, etc. We manually define 21 irrelevant features that are roughly orthogonal to these relevant features (See the full lists in Appendix~\ref{app:feature_list}). The useful training signals come from differences between $y^+$ and $y^-$ that are along \textit{relevant} features and are pointing in the \textit{correct direction} ($y^+$ is \textit{better} than $y^-$ for a \textit{relevant} feature), which we call \textit{desired feature differences}.

\paragraph{Metrics} We define $f_\Delta$ as the number of features along which $y^+$ and $y^-$ differ. To measure the fraction of desired feature differences, we define $f_\Delta^{\mathrm{des}}$ as the fraction of features in $f_\Delta$ that are (a) relevant and (b) contrasted in the correct direction (i.e. $y^+$ is ``better'' than $y^-$ for that feature). Fraction of features that only satisfy condition (a) is denoted by $f_\Delta^{\mathrm{rel}}$. Similar to \texttt{DCRM}, $f_\Delta^{\mathrm{des}}$ indicates the ratio of useful contrast signals among noisy signals.

To compute these, we prompt GPT-4o-mini-0718 to (1) identify the three most prominent features that differ between the two responses (setting $f_\Delta$=$3$) and (2) indicate a contrast direction for each feature if applicable (i.e., whether $y^+$ is better). Referring to the list of relevant features, we can then compute $f_\Delta^{\mathrm{rel}}$ and $f_\Delta^{\mathrm{des}}$. Note that we can use this to study the training dataset (i.e. $Y^+$-$Y^-$), and the learned differences after training ($Y_{\mathrm{trained}}$-$Y_{\mathrm{ref}}$).

\paragraph{Analysis of training datasets ($Y^+-Y^-$)} To study the feature differences LLMs \textit{see} during training, we compute the average $f_\Delta^{\mathrm{rel}}$ and $f_\Delta^{\mathrm{des}}$ across 200 randomly sampled $(y^+, y^-)$ from the training dataset. Higher $f_\Delta^{\mathrm{des}}$ implies higher dataset quality.

\paragraph{Analysis of learning outcomes ($Y_{\mathrm{trained}}-Y_{\mathrm{ref}}$)} To study what LLMs actually \textit{learn} after training, we compute $f_\Delta^{\mathrm{rel}}$ and $f_\Delta^{\mathrm{des}}$ for 200 randomly sampled $(y_{\mathrm{trained}} \sim \pi_{\theta}(x), y_{\mathrm{ref}} \sim \pi_{\mathrm{ref}}(x))$ pairs where $x$ is a test prompt in the AlpacaEval dataset. Higher $f_\Delta^{\mathrm{des}}$ implies that the model learns more useful signals (e.g., to be more helpful) and fewer noisy ones (e.g., to be more sarcastic). 

Following \S~\ref{sec:result} and \S~\ref{sec:implication}, we compare different preference datasets in \S~\ref{sec:feature_analysis}, and then show how $BoN^2$ can improve response pair quality in \S~\ref{sec:new_feature_analysis}.

\subsection{Comparing Common Preference Datasets}
\label{sec:feature_analysis}

We present the results in Table~\ref{tab:feature_analysis} to understand (1) what the model \textit{sees} during training and (2) what it actually \textit{learns}.

\begin{table}
  \small
  \centering
  \begin{tabular}{llcc|cc}
    \toprule
      & & \mc{2}{c}{\textbf{$Y^+$-$Y^-$}} & \mc{2}{c}{\textbf{$Y_{\mathrm{trained}}$-$Y_{\mathrm{ref}}$}} \\
      & & $f_\Delta^{\mathrm{rel}}$ & $f_\Delta^{\mathrm{des}}$ & $f_\Delta^{\mathrm{rel}}$ & $f_\Delta^{\mathrm{des}}$ \\
    \rowcolor{lightgray}\mc{6}{c}{$\pi_{\mathrm{ref}}=$LLaMA2 (\textit{\lmatwos})}\\
    SS-RM & $\pi_{\mathrm{ref}}$ & 63.83 & \textbf{41.83} & 53.75 & \textbf{29.81} \\
     & Gma2      & 56.42 & 38.08 & 53.94 & 29.53 \\
     & Mst     & 62.83 & 37.83 & 54.00 & 29.19 \\
    \midrule
    DS-RM & Gma2-Mst & 61.75 & 39.92 & 53.31 & 28.66 \\
    \midrule
    DS-Fix & Gma2-Mst & 62.5 & 36.33 & 52.22 & 18.83\\
    \rowcolor{lightgray}\mc{6}{c}{$\pi_{\mathrm{ref}}=$LLaMA3.2 (\textit{\lmathro})}\\
    SS-RM & $\pi_{\mathrm{ref}}$ & 64.67 & \textbf{43.25} & 60.08 & 37.50 \\
    & Gma2      & 56.42 & 38.08 & 59.00 & \textbf{37.58} \\
    & Mst     & 62.83 & 37.83 & 61.00 & 35.58 \\
    \midrule
    DS-RM & Gma2-Mst & 61.75 & 39.92 & 60.33 & 34.17 \\
    \midrule
    DS-Fix & Gma2-Mst & 62.50 & 36.33 & 60.17 & 23.33 \\
    \bottomrule
  \end{tabular}
  \caption{\label{tab:feature_analysis}
    $f_\Delta^{\mathrm{des}}$: Percentage of desired feature differences among the identified feature differences; $f_\Delta^{\mathrm{rel}}$: Percentage of relevant feature differences; $Y^+$-$Y^-$: differences identified between $y^+$ and $y^-$ in the training set; $Y_{\mathrm{trained}}$-$Y_{\mathrm{ref}}$: differences identified between model's output on AlpacaEval after training ($Y_{\mathrm{trained}}$) and before training ($Y_{\mathrm{ref}}$). SS-RM datasets typically have the highest $f_\Delta^{\mathrm{des}}$, followed by DS-RM and then DS-Fix.
  % \tg{this table is very confusing becase the Y+ - y- things are same except for pi ref, this should ideally not be listed twice.}\ch{I personally think repeating the numbers twice allows for clearer left to right comparison}
  }
  \vspace{-\baselineskip}
\end{table}

\paragraph{DS-Fix datasets have the lowest proportion of desired feature differences in its training data.} Analyzing the training set $Y^+$-$Y^-$, we see that response pairs from $\pi_{\mathrm{ref}}$ (\texttt{SS-RM}) have the highest percentage of desired feature differences, indicating the highest quality. On the other hand, DS-Fix has the lowest percentage. These results are consistent with our observations in Table~\ref{tab:main_res}. Surprisingly DS-RM has a higher $f_\Delta^{\mathrm{des}}$ than Gma2 (\texttt{SS-RM}) and Mst (\texttt{SS-RM}). A possible explanation will be their actual marginal differences in dataset quality since at least 1 side of the response sources overlap.
% \tg{i dont understand what this mean? marginal differences between what?}

\begin{figure}[t]
    \centering
    \includegraphics[width=0.9\linewidth]{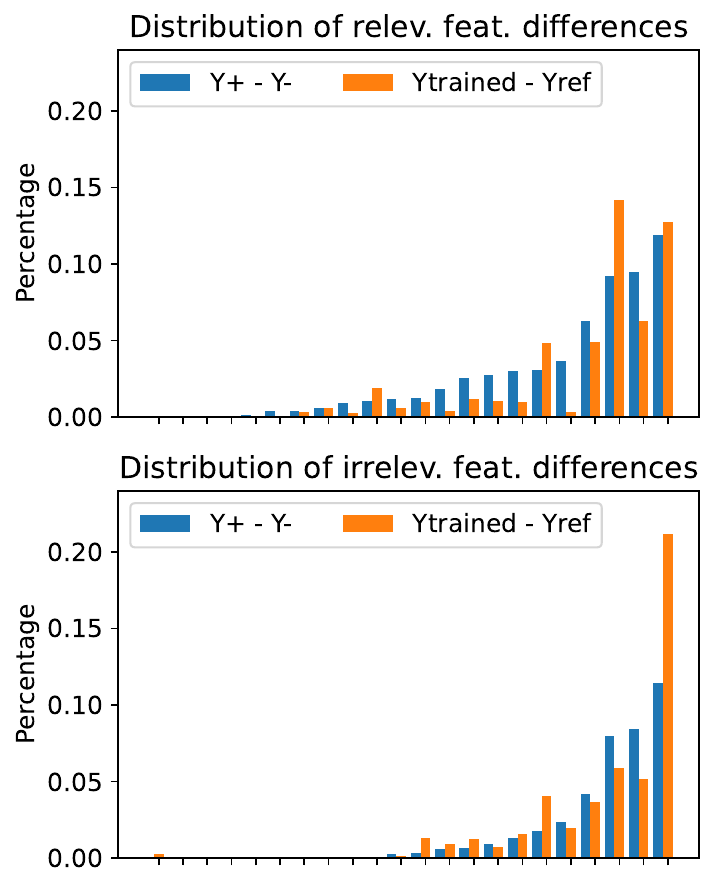}
    \caption{Distributions of relevant (top) and irrelevant (bottom) feature differences. Each pair of adjacent blue and orange bars represents the percentage of a kind of feat. diff. ($y^+$ more helpful, $y^-$ less truthful, etc.) among the identified feat. diff. Blue: training set differences ($Y^+$-$Y^-$); Orange: differences in model outputs on AlpacaEval after or before training ($Y_{\mathrm{trained}}$-$Y_{\mathrm{ref}}$). $Y^+$-$Y^-$ and 
    $Y_{\mathrm{trained}}$-$Y_{\mathrm{ref}}$ have similar distributions.  
    % \tg{what is the x axis here?.}
    }
    \label{fig:distribution_analysis}
    \vspace{-\baselineskip}
\end{figure}

% \begin{figure}[t]
%     \centering
%     \includegraphics[width=\linewidth]{images/top_feature_differences_2.pdf}
%     \caption{Top 5 relevant features with the largest reward score differences in both settings overlap significantly. X-axis: feature; Y-axis: reward score difference per feature when going from $Y^-$ ($Y_{\mathrm{ref}}$) to $Y^+$ ($Y_{\mathrm{trained}}$).}
%     \label{fig:top_features}
% \end{figure}

\paragraph{Desired feature differences learned by the model are proportional to their presence in the training set.} Our initial observation is that higher $f_\Delta^{\mathrm{des}}$ in the training dataset (i.e. $Y^+$-$Y^-$) generally induces higher $f_\Delta^{\mathrm{des}}$ in $Y_{\mathrm{trained}}$-$Y_{\mathrm{ref}}$. This indicates a consistency between the training set and learned outcome for desired feature differences. To analyze this trend in a fine-grained manner and for more general feature differences, we do the following case study in the LLaMA2 $\pi_{\mathrm{ref}}$ (SS-RM) setting.

% \tg{start a new paragraph below, give it a bold header. }

\paragraph{In general, feature differences learned by the model are proportional to their presence in the training set.} We inspect the distribution of feature differences per category (i.e., the percentage of each kind of feat. diff. among all the identified feat. diff.). Figure~\ref{fig:distribution_analysis} shows that for both relevant and irrelevant features, the distributions for $Y^+$-$Y^-$ and $Y_{\mathrm{trained}}$-$Y_{\mathrm{ref}}$ are similar, with a KL divergence of 0.2109 and 0.1284 respectively, so \textbf{more prominent feature differences in the training set are picked up by the model more after training.}\footnote{See Appendix~\ref{app:reward_difference} for more analysis.}

% \tg{i dont know if we should say these KL are equal. I think we shuld point our that } 

% These show that in general, \textbf{the more prominent a feature difference is, the more it will be picked up by the model after training.}. \tg{you should usually not have paragraphs this long ever. break it up.}

\subsection{Effect of Applying Best-of-$N^2$ Pairing}
\label{sec:new_feature_analysis}

\begin{table}[t]
  \small
  \centering

  \begin{tabular}{llcc|cc}
    \toprule
     & & \mc{2}{c}{\textbf{$Y^+$-$Y^-$}} & \mc{2}{c}{\textbf{$Y_{\mathrm{trained}}$-$Y_{\mathrm{ref}}$}} \\
     & & $f_\Delta^{\mathrm{rel}}$ & $f_\Delta^{\mathrm{des}}$ & $f_\Delta^{\mathrm{rel}}$ & $f_\Delta^{\mathrm{des}}$ \\
    \rowcolor{lightgray}\mc{6}{c}{$\pi_{\mathrm{ref}}=$LLaMA2 (\textit{\lmatwos})}\\
    SS-RM & $\pi_{\mathrm{ref}}$ & 63.83 & \textbf{41.83} & 53.75 & 29.81 \\
    & w/ $BoN^2$ & 64.17 & 41.50 & 54.58 & \textbf{31.58} \\
    \midrule
    SS-RM & Mst     & 62.83 & 37.83 & 54.00 & 29.19 \\
    & w/ $BoN^2$     & 66.25 & 
    \textbf{39.08} & 54.08 & \textbf{30.25} \\
    \midrule
    DS-RM & Gma2-Mst & 61.75 & 39.92 & 53.31 & 28.66 \\
    & w/ $BoN^2$ & 62.83 & \textbf{42.75} & 55.67 & \textbf{30.83} \\
    \rowcolor{lightgray}\mc{6}{c}{$\pi_{\mathrm{ref}}=$LLaMA3.2 (\textit{\lmathro})}\\
    SS-RM & $\pi_{\mathrm{ref}}$ & 64.67 & 43.25 & 60.08 & 37.50 \\
    & w/ $BoN^2$ & 65.00 & \textbf{44.83} & 59.67 & \textbf{38.42} \\
    \midrule
    SS-RM & Mst     & 62.83 & 37.83 & 61.00 & \textbf{35.58} \\
    & w/ $BoN^2$     & 65.17 & \textbf{40.25} & 59.33 & 34.25 \\
    \midrule
    DS-RM & Gma2-Mst & 61.75 & 39.92 & 60.33 & 34.17 \\
    & w/ $BoN^2$ & 62.83 & \textbf{41.42} & 60.33 & \textbf{36.75} \\
    \bottomrule
  \end{tabular}
  \caption{\label{tab:new_feature_res}
    Results for feature-based analysis. $BoN^2$ datasets have a higher $f_\Delta^{\mathrm{des}}$ in most settings. 
    % \tg{is any of this improvement statistically significant?}
  }
  \vspace{-\baselineskip}
\end{table}

\paragraph{Best of $N^2$ pairing produces datasets with a higher proportion of desired feature differences.} We conduct the same feature-based analysis as in \S~\ref{sec:feature_analysis}. Table~\ref{tab:new_feature_res} indicates that in most settings, the datasets produced by our method have a higher percentage of desired feature differences (See $f_\Delta^{\mathrm{des}}$ in $Y^+$-$Y^-$), which guides the models to learn effectively and do better in relevant features after training (See $f_\Delta^{\mathrm{des}}$ in $Y_{\mathrm{trained}}$-$Y_{\mathrm{ref}}$). In the LLaMA2 ~$\pi_{\mathrm{ref}}$ (SS-RM) setting, $f_\Delta^{\mathrm{des}}$ in $Y^+$-$Y^-$ remains approximately the same after applying our method, which can be caused by what we discuss in \S~\ref{sec:new_res}.

\section{Related Work}
% \tg{this whole section is extrememely verbose. the language should be made crisper.}\ch{I tried cutting certain content.}

\paragraph{Preference Optimization} Preference Optimization is an alternative to traditional RLHF methods \citep{ouyang2022} such as PPO \citep{schulman2017}. It avoids the need for an explicit reward model. Popular PO algorithms includes DPO \citep{rafailov2024}, IPO \citep{azar2023}, KTO \citep{ethayarajh2024}, R-DPO \citep{park2024}, SimPO \citep{meng2024}, CPO \citep{xu2024a}, ORPO \citep{hong2024}, and so on. Many papers report performance increases on AlpacaEval when training LLMs using PO methods on chat datasets \citep{ding2023, cui2023}.

% Training LLMs with these methods on instructional conversation datasets \citep{ding2023, cui2023} can increase performance on AlpacaEval and MT-Bench \citep{zheng2023}.

\vspace{-2mm}

\paragraph{Response Pairs} 
% PO methods typically require a preferred $y^+$ and dispreferred response $y^-$. 
The choice of response pairs in PO affects training outcomes. \citet{tajwar2024} and \citet{tang2024} investigate response sources and illustrate the benefits of sampling responses on policy. Another line of work focuses on the differences between $y^+$ and $y^-$. 
% If $y^+$ and $y^-$ differ only in desirable feature dimensions, then ideally the trained LLMs should precisely capture the degree to which they differ. 
Prior work \citep{fisch2024, amini2024, furuta2024} suggests that LLMs should learn a different reward margin for each example, since different response pairs can vary in their contrastiveness (i.e., $y^+$ is \textit{much} or \textit{only a little} better than $y^-$).

In reality, however, $y^+$ and $y^-$ often differ in features irrelevant for the task, and a larger gap between them is not always desirable. Certain work focuses on eliminating specific irrelevant differences such as length \citep{singhal2023}. Others take a more general perspective. \citet{wu2024} use reward margins to measure differences and dynamically 
% adjusts the $\beta$ value for each example to 
scales the training signals 
% inverse to the gap
for each example. 
% Examples with large gaps are filtered out. 
\citet{doosterlinck2024} and \citet{guo2024} construct minimally different pairs by revising $y^-$ with a stronger LLM to get $y^+$. However, these methods either do not accurately model the relationship between response pair differences and quality, or require a stronger LLM to be present.

\section{Conclusion}
\vspace{-2mm}
We propose a metric called \texttt{DCRM} that measures the density of useful training signals in response pairs and show its correlation with the PO training outcome. Inspired by this correlation, we design a Best of $N^2$ pairing method, which can curate high-quality datasets to train LLMs with PO effectively. In addition, we provide a feature analysis to inspect the characteristics of various common datasets with varying \texttt{DCRM} values.

% There are many directions worth exploring. Firstly, we use \texttt{DCRM} to select pairs from an existing pool of responses. It is intriguing to know how well our Best of $N^2$ method works as $N$ increases. Secondly, other formulations of \texttt{DCRM} are possible, so long as they encourage small distances and large reward margins.

\section*{Limitations}

Due to time constraints, we only focus on general chat datasets and benchmarks for training and evaluation. While we do provide evaluation results for more task-specific benchmarks such as GSM8K, more efforts can be made to instead train LLMs in these task-specific settings to validate our claims.

% While the intuition behind DCRM is clear, DCRM cannot exactly measure the quality of the response pairs. Firstly, DCRM cannot exactly measure the density of desired differences since it uses distance and reward heuristics to quantify differences. Secondly, there can be other aspects that affect the quality of a response pair, such as the amount, rather than density, of useful signals, which DCRM cannot capture. A more careful and holistic formulation of the metric can be made in future work.

In addition, our $BoN^2$ method works with an existing pool of responses. Instead of having to sample multiple responses per prompt, an alternative to our method will be to use constrained decoding to guide the response generation process toward a high \texttt{DCRM} value. 

% Thirdly, while this work studies the ratio between desired and total differences in response pairs, the amount of these differences themselves may affect PO training effectiveness too. More work can be done in this space too.

% In addition, we only explore our Best of $N^2$ pairing method with $N=5$. Firstly, it is intriguing to know how well our method works as $N$ changes. Secondly, instead of having to sample multiple responses per prompt, an alternative to our method will be to use constrained decoding to guide the response generation process toward a high \texttt{DCRM} value. More work can be done in this space too.

\section*{Ethics Statement}

After manual inspection, we are confident that our work adheres to ethical guidelines. We use Ultrafeedback prompts to curate our datasets, which are open-sourced and publicly available, without the presence of sensitive or private content.

% Bibliography entries for the entire Anthology, followed by custom entries
%\bibliography{anthology,custom}
% Custom bibliography entries only
\bibliography{custom}

\appendix
\section{Preliminary Study in the DS-Fix setting}
\label{app:preliminary}

Although prior work \citep{doosterlinck2024} has shown that sampling responses from different sources gives different performances on chat benchmarks like AlpacaEval \cite{dubois2024}, a missing piece is a qualitative understanding of how the choice of these sources shapes the learned behaviors of LLMs.

In an early pilot study in the DS-Fix setting, we observe a trend for LLMs to over-exploit benign features when $y^+$ and $y^-$ have consistent stylistic differences, which in turn leads to worse performance after training. The following are 2 examples that demonstrate this.

% \paragraph{Setup}: To enforce an explicit stylistic difference, we consider setting xx from above where yl and yw are sampled from different distributions, and the preference label 

\begin{figure*}[h]
    \centering
    \includegraphics[width=1.0\textwidth]{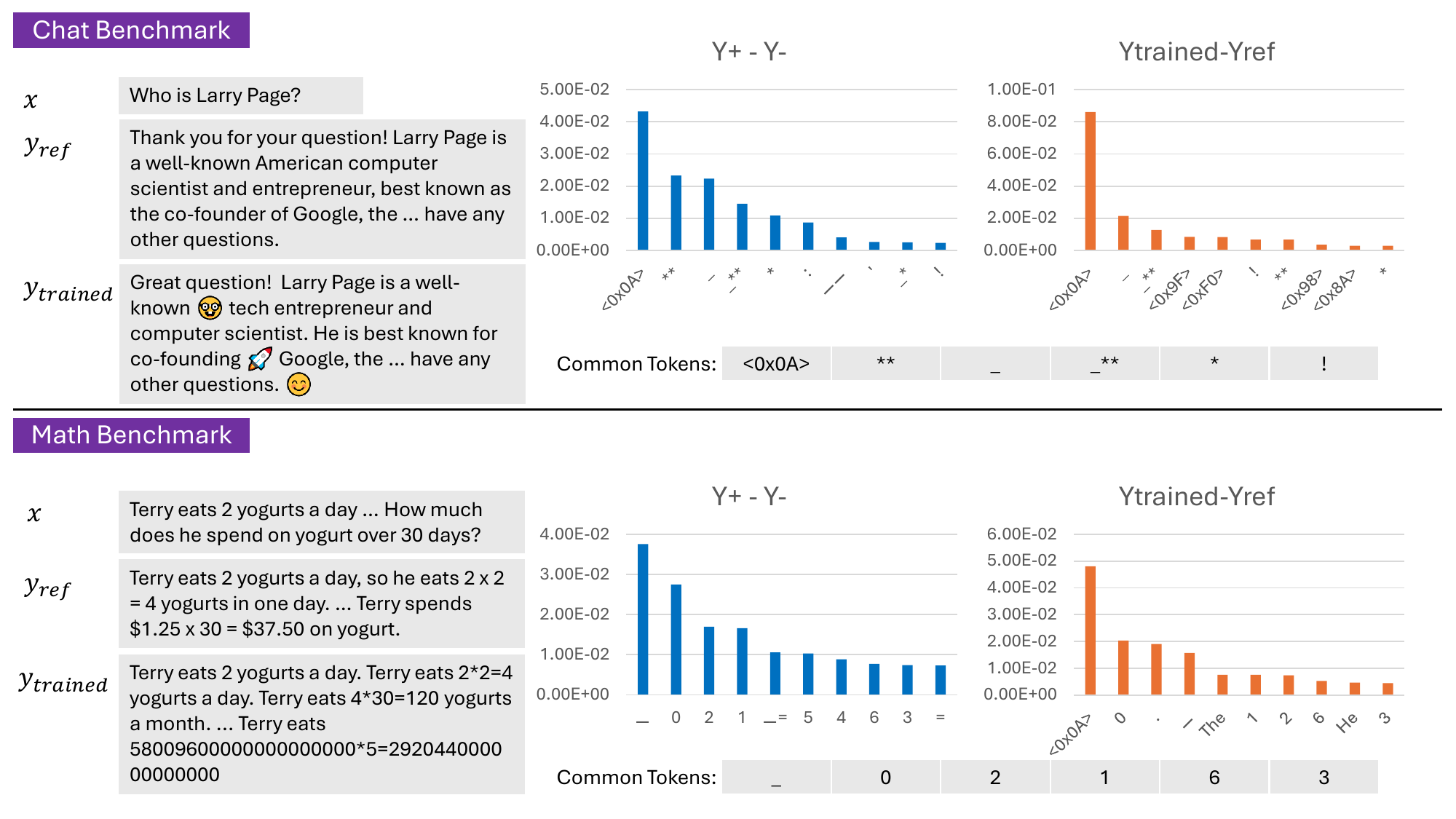}
    \caption{Top: Case Study with Chat Benchmark; Bottom: Case Study with Math benchmark; Left: Example of LLM's output before training ($y_{ref}$) and after training ($y_{trained}$); Middle: Top 10 tokens whose frequency increases the most when changing from $Y^-$ to $Y^+$ in the training set; Right: Top 10 tokens whose frequency increases the most when changing from the model's output before training ($Y_{ref}$) to after training ($Y_{ref}$) on the test set.}
    \label{fig:preliminary_1}
\end{figure*}

\paragraph{Case Study I: Chat Benchmark}

We use the 60K prompts from Ultrafeedback \citep{cui2023} and sample $y^+$ from a strong model \gemtwo~\citep{riviere2024} and $y^-$ from a weak model \mis~\citep{jiang2023}. We set $\pi_{ref}$ to \lmatwos~\citep{touvron2023} and train it with DPO for 2 epochs. We evaluate its performance on AlpacaEval.

\begin{table}[h]
  \small
  \centering
  \begin{tabular}{lccc}
    \toprule
     & \mc{3}{c}{\textbf{AlpacaEval}} \\
     & LC & WR & Length \\
    \midrule
    \lmatwos & 12.57 & 10.43 & 1502 \\
    \midrule
    +Gma2-Mst & 13.26 & 8.99 & 1166 \\
    \bottomrule
  \end{tabular}
  \caption{\label{tab:preliminary_case_chat}
    Result on AlpacaEval. LC: length controlled win rate; WR: raw win rate. The model's raw win rate decreases after training.
  }
\end{table}

Surprisingly, 
% in spite of the fact that both sampling sources are stronger than $\pi_{ref}$, 
the model's raw win rate decreases after training (See Table~\ref{tab:preliminary_case_chat}).
We then closely inspect the model's output. Compared with $\pi_{ref}$, the trained model tends to generate more emojis and other stylistic symbols (See example on the top left of Figure~\ref{fig:preliminary_1}). 

Quantitatively, we conduct a token-level analysis, where we calculate the average frequency for each token to appear in models' responses to AlpacaEval questions before training ($Y_{ref}$) or after training ($Y_{trained}$) (See details in Appendix~\ref{app:token_analysis}). We then check the tokens whose frequency increases the most when going from $Y_{ref}$ to $Y_{trained}$ (See Figure~\ref{fig:preliminary_1} top right). As expected, 5 out of the top 10 tokens are emoji tokens (those surrounded by <>). The rest are mostly also stylistic tokens (** and * are used to bold text and create bullet points).

These stylistic features are indeed learned from the training set. We calculate the same frequency differences for each token when changing from dispreferred responses $Y^-$ to preferred responses $Y^+$, and found the same emoji token (<0x0A>) and other stylistic tokens (**, *, etc.) to appear much more frequently in $Y^+$ than in $Y^-$.

\paragraph{Case Study II: Math Benchmark}

We also conduct experiments on a Math Benchmark, GSM8K \cite{cobbe2021}. We adopt the setting from SPIN \cite{chen2024} and set $y^+$ to be the responses from human annotators and $y^-$ to be the responses from $\pi_{ref}$ (\lmatwos). We then use DPO to train $\pi_{ref}$ for 5 epochs, on 6,725 examples from GSM8K's original training split. We use the remaining 748 examples for validation and select the best checkpoint. Similar to the previous case study, we again observe a surprising performance drop on GSM8K's test split.

\begin{table}[h]
  \small
  \centering
  \begin{tabular}{lc}
    \toprule
     & GSM8K ACC (0-shot) \\
    \midrule
    \lmatwos & 23.88 \\
    \midrule
    +Human-$\pi_{ref}$ & 18.20 \\
    \bottomrule
  \end{tabular}
  \caption{\label{table:preliminary_case_math}
    Result on GSM8K; ACC: Accuracy; The model's accuracy decreases after training.
  }
\end{table}

Manual inspection suggests that the model tends to generate repetitive sentences that include nonsensical math calculations (Figure~\ref{fig:preliminary_1} bottom left). The token-level analysis reveals that the model learns to generate more digits, which is also attributable to the difference between $Y^+$ and $Y^-$ in the training set (Figure~\ref{fig:preliminary_1} bottom right and middle).

The above suggests that differences between $y^+$ and $y^-$ in irrelevant spurious features in the training set cause LLMs to pick up these features instead of those targeted ones (correctness, etc.). This leads us to hypothesize that when the \textit{proportion} (or \textit{density}) of truly useful contrast signals decreases among all the contrast signals in the response pair, training becomes less effective.

\subsection{Token-level Analysis}
\label{app:token_analysis}
We define a length normalized bag of words representation of a sequence $y$ as follows: we count for each token $t$ in the vocabulary $V$ its number of occurrences in $y$, which we denote as $n(t, y)$. We then divide it by the length of $y$, $|y|$, to get $bow_n(t, y) = \frac{n(t, y)}{|y|}$. This tells how much of $y$ is made up of $t$. We then compute the average of this value across the model's responses to AlpacaEval queries after training ($Y_{trained}$) to get $bow_n(t, Y_{trained}) = \frac{\sum_{y\in Y_{trained}}{bow_n(t, y)}}{|Y_{trained}|}$, and similarly $bow_n(t, Y_{ref})$ for model's responses before training ($Y_{ref}$).

The difference between $bow_n(t, Y_{trained})$ and $bow_n(t, Y_{ref})$ tells how much more frequently $t$ appears in the model's responses after training. Similarly, we can take the preferred responses $Y^+$ and dispreferred responses $Y^-$ in the training set, and search for tokens that occur more frequently in $Y^+$.

\section{Training Details}
\label{app:training_details}
We set $\beta=0.1$, and train the model for 2 epochs. We use Adam Optimizer with a learning rate of 5e-7, warmup ratio of 0.1, and a cosine learning schedule.

\section{Feature Difference Analysis}
\label{app:feature_difference_analysis}

\subsection{Relevant and Irrelevant Features}
\label{app:feature_list}
\begin{table}[!htp]
    \begin{tabular}{p{0.9\linewidth}}
        \toprule
          Relevant Features \\
        \midrule
         "helpfulness", "correctness", "factuality", "coherence", "verbosity", "instruction following", "truthfulness", "honesty", "harmlessness", "code complexity", "code readability" \\
        \midrule
         Irrelevant Features \\
        \midrule
        "writing style", "tone", "politeness", "friendliness", "caring or not", "intimacy", "empathy", "language type", "casual or formal", "authoritative or not", "creativity", "certainty", "humor", "passive or active", "pessimistic or optimistic", "explicit or implicit", "sarcastic or not", "passion", "repetitiveness", "word usage diversity", "structure of presentation", "other" \\
        \bottomrule
    \end{tabular}
    \caption{\label{tab:features} Complete List of Relevant and Irrelevant Features}
\end{table}

We define the relevant features to be the 11 features synthesized from the 19 reward features modeled by ArmoRM. As for the irrelevant features, we manually select 21 features that are not directly related to the relevant features and include an additional "other" feature that refers to all other features not specified in the list. See details in Table~\ref{tab:features}.

\subsection{Prompt}
\label{app:feature_prompt}
The prompt is shown in Table~\ref{tab:sequence_analysis_prompt}. We instruct the judge to identify the top 3 features in which the 2 given responses differ, and the corresponding contrast directions if applicable. To avoid potential biases, we do not reveal the source of each response ($y^+$ or $y^-$; $y_{trained}$ or $y_{ref}$). Additionally, we ask the judge to give 2 separate predictions where in the first prediction $y_1=y^+(y_{trained}), y_2=y^-(y_{ref})$ and in the second prediction $y_1=y^-(y_{ref}), y_2=y^+(y_{trained})$, respectively.

\begin{table*}[!htp]
    \rule{\textwidth}{1pt}
Given 2 responses y1 and y2 to a query x, identify the top 3 most prominent features in which y1 and y2 differ. Provide a justification for each feature that you identified. The features that you identified should only come from the following set of potential features:\\

\{explicit or implicit, instruction following, code readability, caring or not, pessimistic or optimistic, writing style, certainty, truthfulness, casual or formal, tone, intimacy, code complexity, passion, friendliness, passive or active, authoritative or not, word usage diversity, correctness, politeness, language type, factuality, empathy, creativity, coherence, repetitiveness, verbosity, sarcastic or not, structure of presentation, harmlessness, humor, helpfulness, honesty\}\\

Note that the features "code complexity" and "code readability" are only applicable for programming or coding tasks. Do not indicate these for non programming or coding tasks.\\

If you think none of the feature listed above can explain the differences between y1 and y2, propose new features that can explain the differences. Again, provide a justification for each proposed new feature.\\

Additionally, for any feature where it makes sense to say y1 is "better" or "worse" than y2 in terms of that feature (e.g., helpfulness, where more helpful is better; verbosity, where less verbose is better), identify which response is better. You should put "y1" or "y2". For other features where differences do not imply "better" or "worse" (writing style, tone, formal or casual, language type, etc.), put "Not applicable".\\

Give your response in the following JSON format:\\

\{\\
    \indent\hspace{1cm}feature 1: \{\\
        \indent\hspace{2cm}"justification": justification 1,\\
        \indent\hspace{2cm}"better response": "y1" or "y2" or "Not applicable"\\
    \indent\hspace{1cm}\},\\
    \indent\hspace{1cm}...\\
    \indent\hspace{1cm}feature 3: \{\\
        \indent\hspace{2cm}"justification": justification 3,\\
        \indent\hspace{2cm}"better response": "y1" or "y2" or "Not applicable"\\
\indent\hspace{1cm}\}\\
\}\\

Query x: \{x\}\\\\

Response y1: \{y1\}\\\\

Response y2: \{y2\}\\\\

Answer:\\
    \rule{\textwidth}{1pt}
    \caption{\label{tab:sequence_analysis_prompt} Prompt for Sequence-level Analysis.}
\end{table*}

\subsection{Reward differences for relevant features}
\label{app:reward_difference}
\begin{figure}[t]
    \centering
    \resizebox{0.479\textwidth}{!}{
    \includegraphics[width=\linewidth]{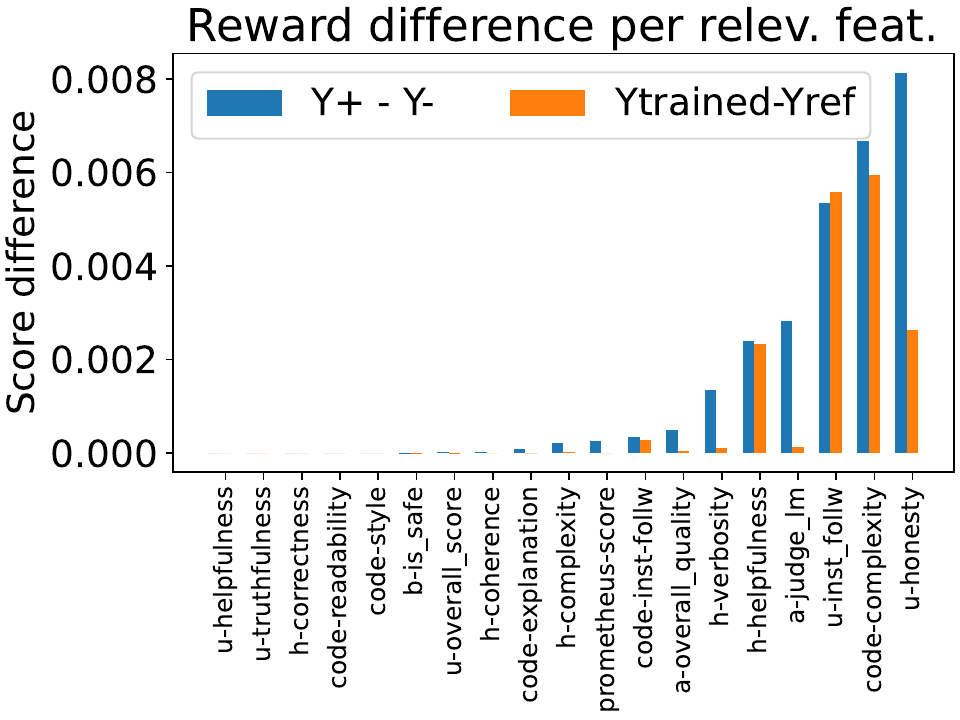}}
    \caption{The fine-grained, per feature reward score differences in both settings overlap significantly. X-axis: relevant feature. u: Ultrafeedback, h: Helpsteer \citep{wang2024b}, a: Argilla, b: BeaverTails \citep{ji2023}; Y-axis: reward score difference per feature when going from $Y^-$ ($Y_{\mathrm{ref}}$) to $Y^+$ ($Y_{\mathrm{trained}}$).}
    \label{fig:rm_score_per_feature}
\end{figure}

\paragraph{Reward differences of relevant features follow similar distributions between $Y^+$-$Y^-$ and $Y_{\mathrm{trained}}$-$Y_{\mathrm{ref}}$.} Since we have the fine-grained reward score for each of the relevant features from ArmoRM\footnote{These are the 19 original, unsynthesized features, containing duplications.}, we compute the change in reward score per feature. Consistent with what we notice in \S~\ref{sec:feature_analysis}, Figure~\ref{fig:rm_score_per_feature} shows that the reward score changes in $Y^+$-$Y^-$ and $Y_{\mathrm{trained}}$-$Y_{\mathrm{ref}}$ are similar. In particular, the top 3 features with the highest changes, which explain over 50 percent of the total reward score changes, are the same for both settings (i.e., the top 3 are honesty, code complexity, and instruction following in both settings).

% \subsection{Qualitative Analysis of Feature Difference Identification Precision}
% \label{app:feature_precision}
% To check how accurate the judge model is, we randomly select 30 examples from the Gma2-Mst (DS-Fix) dataset and inspect the precision of the judge model in identifying the most prominent features. We manually inspect each of the 6 identified feature differences per example (the judge goes through each example twice), and count the number of feature differences that we think are truly among the top 3 most prominent feature differences for that example. We find that 165 of the 180 feature differences are truly the most prominent, giving an estimated identification precision of 0.9167 for the judge.

\section{\texttt{DCRM} Properties}
\label{app:dcrm_properties}
Our \texttt{DCRM} metric has the following properties.

\paragraph{1. Encourage high reward margin, low distance.} Denote the distance $e_\Delta+p_\Delta$ as d. For any response pairs $p_{ij}$ and $p_{i'j'}$, if $r_\Delta(p_{ij}) > r_\Delta(p_{i'j'})$ and $d(p_{ij}) = d(p_{i'j'})$, then $DCRM(p_{ij}) > DCRM(p_{i'j'})$. Similarly, if $r_\Delta(p_{ij}) = r_\Delta(p_{i'j'})$ and $d(p_{ij}) < d(p_{i'j'})$, then $DCRM(p_{ij}) > DCRM(p_{i'j'})$.

\paragraph{2. Preserve reward margin sign.} \texttt{DCRM} always has the same sign as the reward margin. 
% any response pair with a positive reward margin has a higher \texttt{DCRM} value than any pair with a 0 reward margin, which in turn has a higher \texttt{DCRM} value than any pair with a negative reward margin
For any pairs $p_{ij}$, $p_{i'j'}$, $p_{i''j''}$ where $r_\Delta(p_{ij})<0$, $r_\Delta(p_{i'j'})=0$, and $r_\Delta(p_{i''j''})>0$, we should have $DCRM(p_{i''j''})>DCRM(p_{i'j'})>DCRM(p_{ij})$. This means any pair with a positive overall training signal has a higher \texttt{DCRM} value than those with an overall neutral signal, followed by those with an overall negative signal. Additionally, for any pairs $p_{ij}$ and $p_{i'j'}$ where $r_\Delta(p_{ij}) = r_\Delta(p_{i'j'}) = 0$, we have $DCRM(p_{ij}) = DCRM(p_{i'j'})$. This means any pairs with an overall neutral training signal have the same \texttt{DCRM} value.

\section{Complete Results}
\label{app:complete_results}

\begin{table*}
  \small
  \centering
  \begin{tabular}{llcc|c|c|cccc}
    \toprule
    & & \mc{4}{c}{\textbf{Performance}} & \mc{4}{c}{\textbf{Dataset Statistics}} \\
    & & AP-L & AP-R & MT & AH & $e_\Delta$ & $p_\Delta$ & $r_\Delta$(e-2) & \texttt{DCRM}(e-2)\\
    \midrule
    & \lmatwos & 12.57 & 10.43 & 5.41 & 8.90 & - & - & - & - \\
    \midrule
    SS-RM & +$\pi_{\mathrm{ref}}$ & 22.36 & 16.81 & 5.55 & 16.67 & 427 & 32.48 & 2.82 & 4.54 \\
    & w/ $BoN^2$ & 22.41 & 17.2 & 5.67 & 16.07 & 370 & 23.87 & 2.52 & 5.94 \\
    & +Mst     &  15.49 & 12.07 & 5.40 & 10.42 & 526 & 158.54 & 2.13 & 1.59 \\
    & w/ $BoN^2$     &  17.42 & 13.29 & 5.48 & 10.99 & 410 & 79.94 & 1.79 & 2.07 \\
    & +Lma3     &  19.59 & 15.49 & 5.38 & 12.62 & 427 & 74.07 & 2.01 & 1.82 \\
    & +Gma2     &  15.89 & 13.12 & 5.50 & 11.57 & 370 & 91.78 & 1.70 & 2.87 \\
    \midrule
    DS-RM & +Gma2-Mst & 14.13 & 11.51 & 5.52 & 10.55 & 542 & 226.47 & 2.03 & 1.13 \\
    & w/ $BoN^2$ ($r_\Delta$ only) & 16.20 & 13.17 & 5.48 & 11.98 & 495 & 257.84 & 3.78 & 2.21 \\
    & w/ $BoN^2$ & 16.82 & 13.6 & 5.48 & 11.75 & 458 & 142.94 & 3.27 & 2.58 \\
    \midrule
    DS-Fix & +Gma2-Lma3 & 14.02 & 9.53 & 5.63 & 10.62 & 490 & 212.21 & 2.21 & 2.08 \\
    & +Gma2-Mst & 13.26 & 8.99 & 5.24 & 6.68 & 542 & 226.47 & 1.02 & 0.43 \\
    \bottomrule
  \end{tabular}
  \caption{\label{tab:complete_results_llama2}
    Results on \lmatwos. Lma3: \lmathre
  }
\end{table*}

% \begin{table*}
%   \small
%   \centering
%   \begin{tabular}{llcc|cc|cccc}
%     \toprule
%     & & \mc{4}{c}{\textbf{AlpacaEval}} & \mc{4}{c}{\textbf{Dataset Statistics}} \\
%     & & LC & WR & LC$_{\mathrm{ref}}$ & WR$_{\mathrm{ref}}$ & $e_\Delta$ & $p_\Delta$ & $r_\Delta$(e-2) & \texttt{DCRM}(e-2)\\
%     \midrule
%     & \lmatwos & 12.57 & 10.43 & 50 & 50 & - & - & - & - \\
%     \midrule
%     SS-RM & +$\pi_{\mathrm{ref}}$ & 22.36 & 16.81 & 71.66 & 70.06 & 427 & 32.48 & 2.82 & 4.54 \\
%     & w/ $BoN^2$ & 22.41 & 17.2 & 72.92 & 72.30 & 370 & 23.87 & 2.52 & 5.94 \\
%     & +Mst     &  15.49 & 12.07 & 60.86 & 58.22 & 526 & 158.54 & 2.13 & 1.59 \\
%     & w/ $BoN^2$     &  17.42 & 13.29 & 59.18 & 57.08 & 410 & 79.94 & 1.79 & 2.07 \\
%     & +Lma3     &  19.59 & 15.49 & 67.37 & 66.25 & 427 & 74.07 & 2.01 & 1.82 \\
%     & +Gma2     &  15.89 & 13.12 & 58.40 & 58.95 & 370 & 91.78 & 1.70 & 2.87 \\
%     \midrule
%     DS-RM & +Gma2-Mst & 14.13 & 11.51 & 56.29 & 56.15 & 542 & 226.47 & 2.03 & 1.13 \\
%     & w/ $BoN^2$ ($r_\Delta$ only) & 16.20 & 13.17 & 59.80 & 59.32 & 495 & 257.84 & 3.78 & 2.21 \\
%     & w/ $BoN^2$ & 16.82 & 13.6 & 61.45 & 60.93 & 458 & 142.94 & 3.27 & 2.58 \\
%     \midrule
%     DS-Fix & +Gma2-Lma3 & 14.02 & 9.53 & 52.42 & 44.43 & 490 & 212.21 & 2.21 & 2.08 \\
%     & +Gma2-Mst & 13.26 & 8.99 & 49.72 & 42.67 & 542 & 226.47 & 1.02 & 0.43 \\
%     \bottomrule
%   \end{tabular}
%   \caption{\label{tab:complete_results_llama2}
%     Results on \lmatwos. Lma3: \lmathre
%   }
% \end{table*}

\begin{table*}
  \small
  \centering
  \begin{tabular}{llcc|c|c|cccc}
    \toprule
    & & \mc{4}{c}{\textbf{Performance}} & \mc{4}{c}{\textbf{Dataset Statistics}} \\
    & & AP-L & AP-R & MT & AH & $e_\Delta$ & $p_\Delta$ & $r_\Delta$(e-2) & \texttt{DCRM}(e-2)\\
    \midrule
    & Gemma-2B-IT    & 16.07 & 10.31 & 4.80 & 5.40 & - & - & - & - \\
    \midrule
    SS-RM & +$\pi_{\mathrm{ref}}$          & 27.03 & 18.01 & 4.97 & 9.58 & 229 & 56.48 & 4.15 & 11.11 \\
    & w/ $BoN^2$     & 28.08 & 17.64 & 4.93 & 10.50 & 197 & 35.93 & 3.74 & 14.90 \\
    & +Mst              & 22.96 & 14.66 & 5.02 & 8.39 & 526 & 244.81 & 2.13 & 1.50 \\
    & w/ $BoN^2$         & 26.71 & 16.89 & 5.03 & 9.58 & 342 & 99.29 & 1.74 & 2.22 \\
    & +Lma3               & 25.49 & 17.04 & 5.15 & 8.63 & 427 & 110.00 & 2.01 & 3.07 \\
    & +Gma2               & 25.13 & 17.76 & 5.19 & 10.22 & 370 & 103.15 & 1.70 & 2.85\\
    % \midrule
    % +Gma2-Mst(DSR)      & 24.81 & 17.89 & 60.98 & 67.02 & 542 & 303.69 & 2.026e-2 & 1.055e-2 \\
    % +Gma2-Mst(DSR+$BoN^2$) & 25.92 & 18.14 & 65.26 & 70.31 & 380 & 169.84 & 3.207e-2 & 2.752e-2 \\
    \midrule
    DS-RM & +Lma3-Mst      & 22.36 & 15.03 & 4.96 & 7.70 & 466 & 295.38 & 1.77 & 1.10 \\
    & w/ $BoN^2$ ($r_\Delta$ only) & 24.41 & 15.16 & 4.98 & 7.09 & 515 & 355.66 & 3.59 & 2.03 \\
    & w/ $BoN^2$ & 26.14 & 17.76 & 5.09 & 9.21 & 393 & 170.61 & 3.03 & 2.80 \\
    \midrule
    % +Gma2-Lma3(DSS)      & 25.52 & 18.14 & 60.16 & 62.36 & 490 & 165.14 & 2.21e-2 & 2.076e-2 \\
    % +Gma2-Mst(DSS)     & 23.39 & 18.01 & 54.31 & 62.17 & 542 & 303.69 & 1.016e-2 & 3.921e-3 \\
    DS-Fix & +Lma3-Mst     & 16.81 & 16.15 & 4.53 & 6.23 & 466 & 295.38 & 0.71 & 0.32 \\
    \bottomrule
  \end{tabular}
  \caption{\label{tab:complete_results_gemma}
    Results on Gemma-2B-IT. Note that for symmetrical purposes, we include an additional Lma3-Mst (DS-RM/DS-Fix) setting in place of the Gma2-Mst (DS-RM/DS-Fix) setting since Gemma and Gma2 are from the same series.
  }
\end{table*}

\begin{table*}
  \small
  \centering
  \begin{tabular}{llcc|c|c|cccc}
    \toprule
    & & \mc{4}{c}{\textbf{Performance}} & \mc{4}{c}{\textbf{Dataset Statistics}} \\
    & & AP-L & AP-R & MT & AH & $e_\Delta$ & $p_\Delta$ & $r_\Delta$(e-2) & \texttt{DCRM}(e-2)\\
    \midrule
    & \lmathro & 14.15 & 15.34 & 4.66 & 10.88 & - & - & - & - \\
    \midrule
    SS-RM & +$\pi_{\mathrm{ref}}$          & 22.80 & 25.65 & 5.01 & 18.88 & 434 & 120.07 & 4.22 & 7.53 \\
    & w/ $BoN^2$     & 24.77 & 27.64 & 5.10 & 20.25 & 356 & 63.55 & 3.58 & 11.48 \\
    & +Mst              & 19.43 & 19.94 & 4.91 & 16.03 & 526 & 176.22 & 2.13 & 1.68 \\
    & w/ $BoN^2$         & 21.73 & 21.37 & 5.11 & 16.62 & 339 & 78.81 & 1.78 & 2.44 \\
    & +Lma3               & 27.81 & 32.73 & 5.16 & 19.35 & 427 & 61.33 & 2.01 & 3.81 \\
    & +Gma2               & 24.57 & 27.52 & 4.99 & 15.91 & 370 & 84.78 & 1.70 & 3.15 \\
    \midrule
    DS-RM & +Gma2-Mst      & 20.01 & 21.61 & 5.01 & 13.61 & 542 & 228.22 & 2.03 & 1.17 \\
    & w/ $BoN^2$ ($r_\Delta$ only) & 21.72 & 24.66 & 4.99 & 17.84 & 495 & 269.60 & 3.78 & 3.02 \\
    & w/ $BoN^2$ & 24.53 & 27.76 & 5.04 & 19.17 & 374 & 134.94 & 3.24 & 3.02 \\
    \midrule
    DS-Fix & +Gma2-Lma3       & 10.31 & 8.20 & 4.84 & 13.23 & 490 & 211.42 & 2.21 & 2.27 \\
    & +Gma2-Mst      & 17.79 & 13.79 & 4.54 & 14.94 & 542 & 228.22 & 1.02 & 0.44 \\
    \bottomrule
  \end{tabular}
  \caption{\label{tab:complete_results_llama32}
    Results on \lmathro
  }
\end{table*}

% \begin{table*}
%   \small
%   \centering
%   \begin{tabular}{llcc|cc|cccc}
%     \toprule
%     & & \mc{4}{c}{\textbf{AlpacaEval}} & \mc{4}{c}{\textbf{Dataset Statistics}} \\
%     & & LC & WR & LC$_{\mathrm{ref}}$ & WR$_{\mathrm{ref}}$ & $e_\Delta$ & $p_\Delta$ & $r_\Delta$(e-2) & \texttt{DCRM}(e-2)\\
%     \midrule
%     & \lmathro & 14.15 & 15.34 & 50 & 50 & - & - & - & - \\
%     \midrule
%     SS-RM & +$\pi_{\mathrm{ref}}$          & 22.80 & 25.65 & 67.89 & 70.00 & 434 & 120.07 & 4.22 & 7.53 \\
%     & w/ $BoN^2$     & 24.77 & 27.64 & 69.11 & 71.49 & 356 & 63.55 & 3.58 & 11.48 \\
%     & +Mst              & 19.43 & 19.94 & 61.29 & 60.06 & 526 & 176.22 & 2.13 & 1.68 \\
%     & w/ $BoN^2$         & 21.73 & 21.37 & 61.70 & 59.25 & 339 & 78.81 & 1.78 & 2.44 \\
%     & +Lma3               & 27.81 & 32.73 & 75.47 & 77.45 & 427 & 61.33 & 2.01 & 3.81 \\
%     & +Gma2               & 24.57 & 27.52 & 67.09 & 69.44 & 370 & 84.78 & 1.70 & 3.15 \\
%     \midrule
%     DS-RM & +Gma2-Mst      & 20.01 & 21.61 & 56.92 & 57.39 & 542 & 228.22 & 2.03 & 1.17 \\
%     & w/ $BoN^2$ ($r_\Delta$ only) & 21.72 & 24.66 & 65.65 & 67.14 & 495 & 269.60 & 3.78 & 3.02 \\
%     & w/ $BoN^2$ & 24.53 & 27.76 & 68.40 & 69.38 & 374 & 134.94 & 3.24 & 3.02 \\
%     \midrule
%     DS-Fix & +Gma2-Lma3       & 10.31 & 8.20 & 34.37 & 26.71 & 490 & 211.42 & 2.21 & 2.27 \\
%     & +Gma2-Mst      & 17.79 & 13.79 & 45.78 & 36.02 & 542 & 228.22 & 1.02 & 0.44 \\
%     \bottomrule
%   \end{tabular}
%   \caption{\label{tab:complete_results_llama32}
%     Results on \lmathro
%   }
% \end{table*}

Table~\ref{tab:complete_results_llama2}, \ref{tab:complete_results_gemma}, \ref{tab:complete_results_llama32} show the complete results and dataset statistics for each $\pi_{\mathrm{ref}}$ that we have trained.
% For each experiment on \lmatwos, we have 3 independent training runs with different seeds and report the average results.

Similar to the main results in \S~\ref{sec:new_res}, we observe that SS-RM generally performs the best and DS-Fix generally performs the worst, and that there is a positive correlation between the average \texttt{DCRM} value of the training dataset and the model's performance boost after training.
% AlpacaEval:
% LLAMA-2-7B-Chat
% GEMMA
% LLAMA32

\section{Correlation Analysis on AlpacaEval}
\label{app:correlation_analysis}

In addition to the 3 SS-RM, 1 DS-RM, and 1 DS-Fix settings discussed in \S~\ref{sec:experiment_setup}, we also include the 8 additional settings for more accurate computation of the correlation. These include the 3 settings in \S~\ref{sec:implication} where we apply our Best of $N^2$ method with \texttt{DCRM} and 5 settings from the ablation study ($e_\Delta$ only, $p_\Delta$ only, $e_\Delta$+$p_\Delta$, $e_\Delta$+$r_\Delta$, $p_\Delta$+$r_\Delta$). 

\subsection{Correlation with individual metrics}
\label{app:corr_individual_metrics}

\begin{figure}[h]
    \centering
    \includegraphics[width=\linewidth]{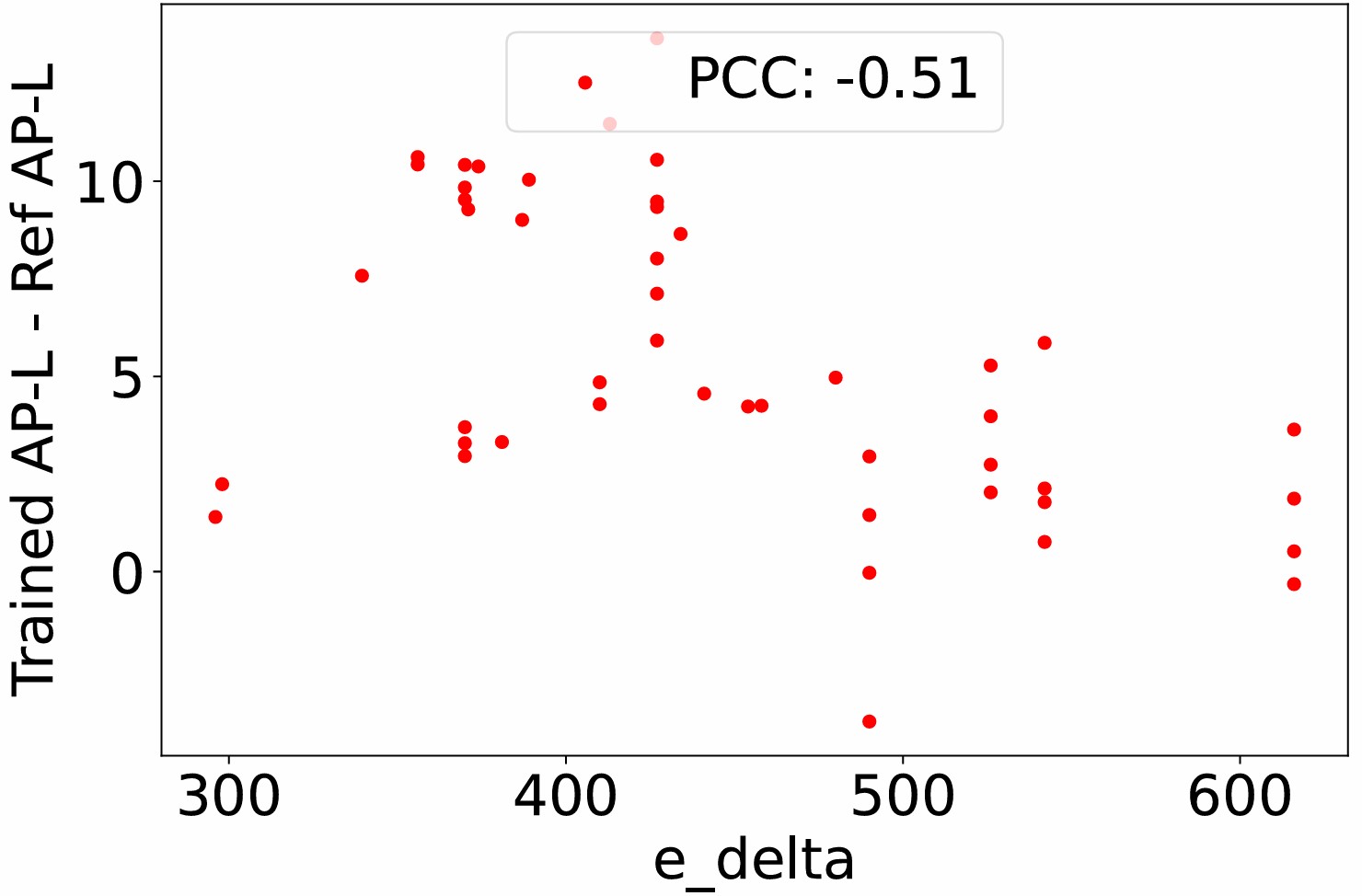}
    \caption{Correlation with $e_\Delta$}
    \label{fig:corr_ted}
\end{figure}

\begin{figure}[h]
    \centering
    \includegraphics[width=\linewidth]{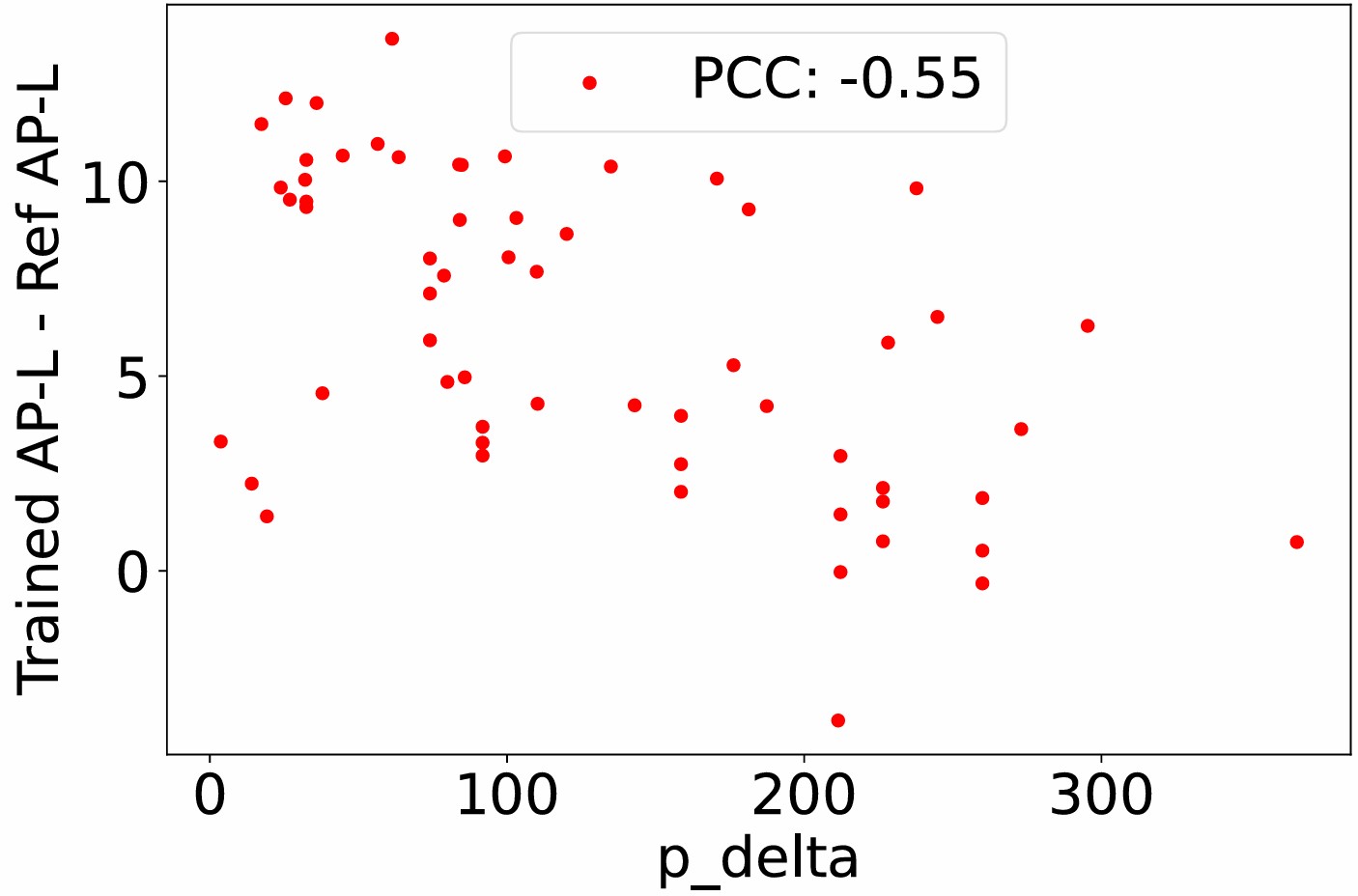}
    \caption{Correlation with $p_\Delta$}
    \label{fig:corr_apd}
\end{figure}

\begin{figure}[h]
    \centering
    \includegraphics[width=\linewidth]{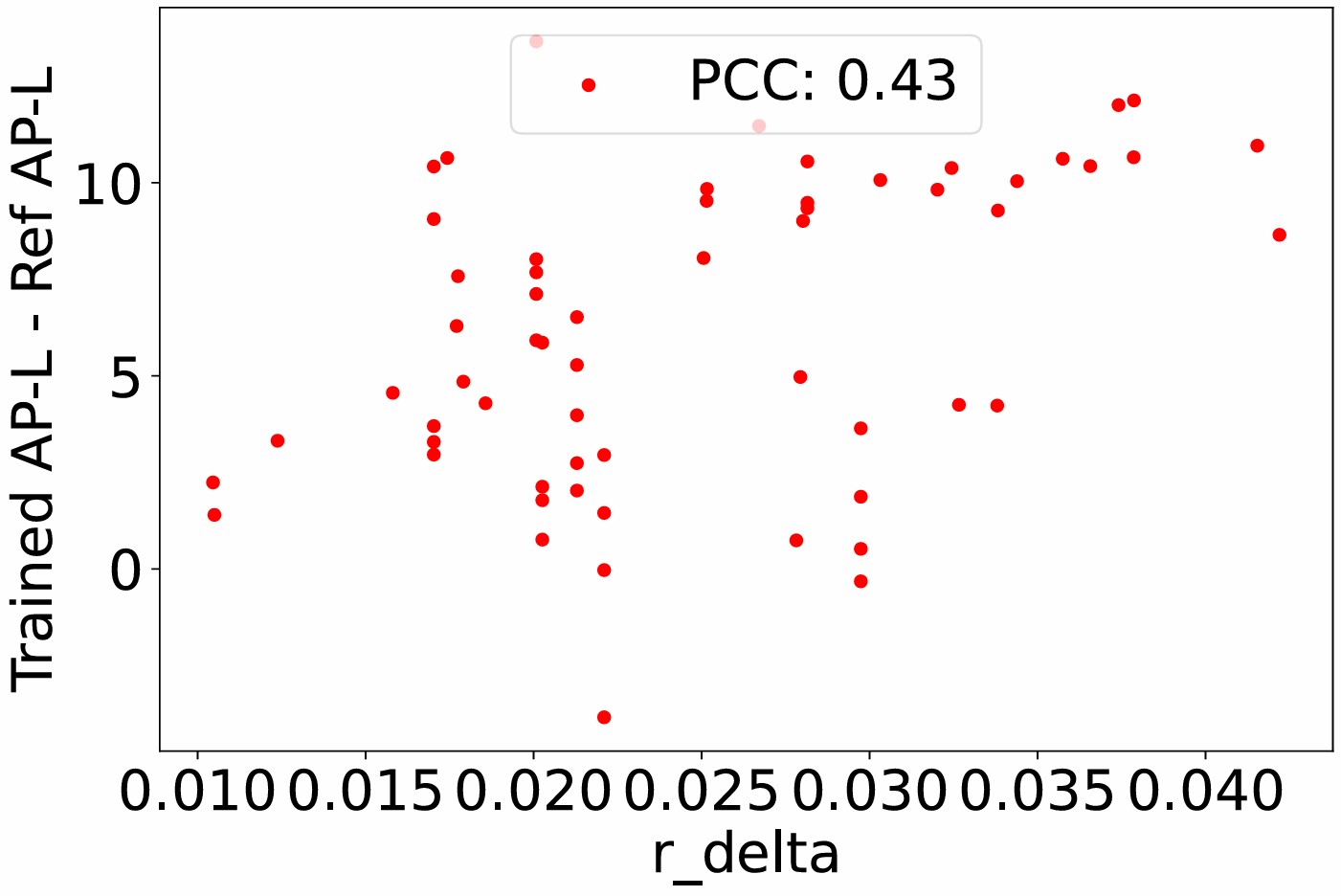}
    \caption{Correlation with $r_\Delta$}
    \label{fig:corr_rw_margin}
\end{figure}

In Figure~\ref{fig:corr_ted}, \ref{fig:corr_apd}, and \ref{fig:corr_rw_margin}, for each individual component of \texttt{DCRM} ($e_\Delta$, $p_\Delta$, and $r_\Delta$), we show the correlation between the training set's metric value and the change in the model's length controlled win rate on AlpacaEval post-training. \texttt{DCRM} has a stronger correlation than these individual metrics.

\section{Correlation with MT-Bench and Arena-Hard}
\label{app:corr_mt_ah}

\begin{figure}[h]
    \centering
    \includegraphics[width=\linewidth]{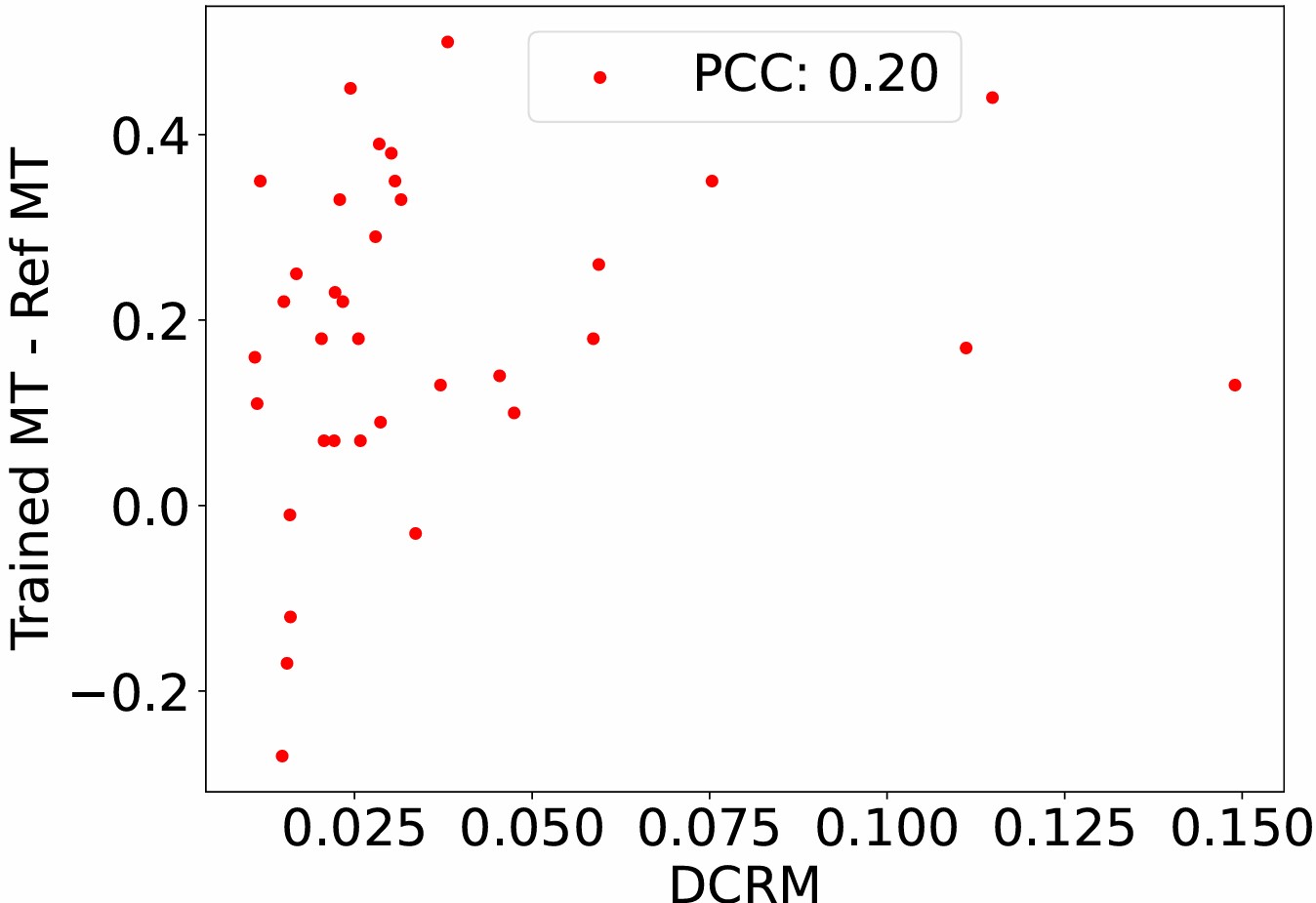}
    \caption{Correlation with MT-Bench Performance}
    \label{fig:corr_mt}
\end{figure}

\begin{figure}[h]
    \centering
    \includegraphics[width=\linewidth]{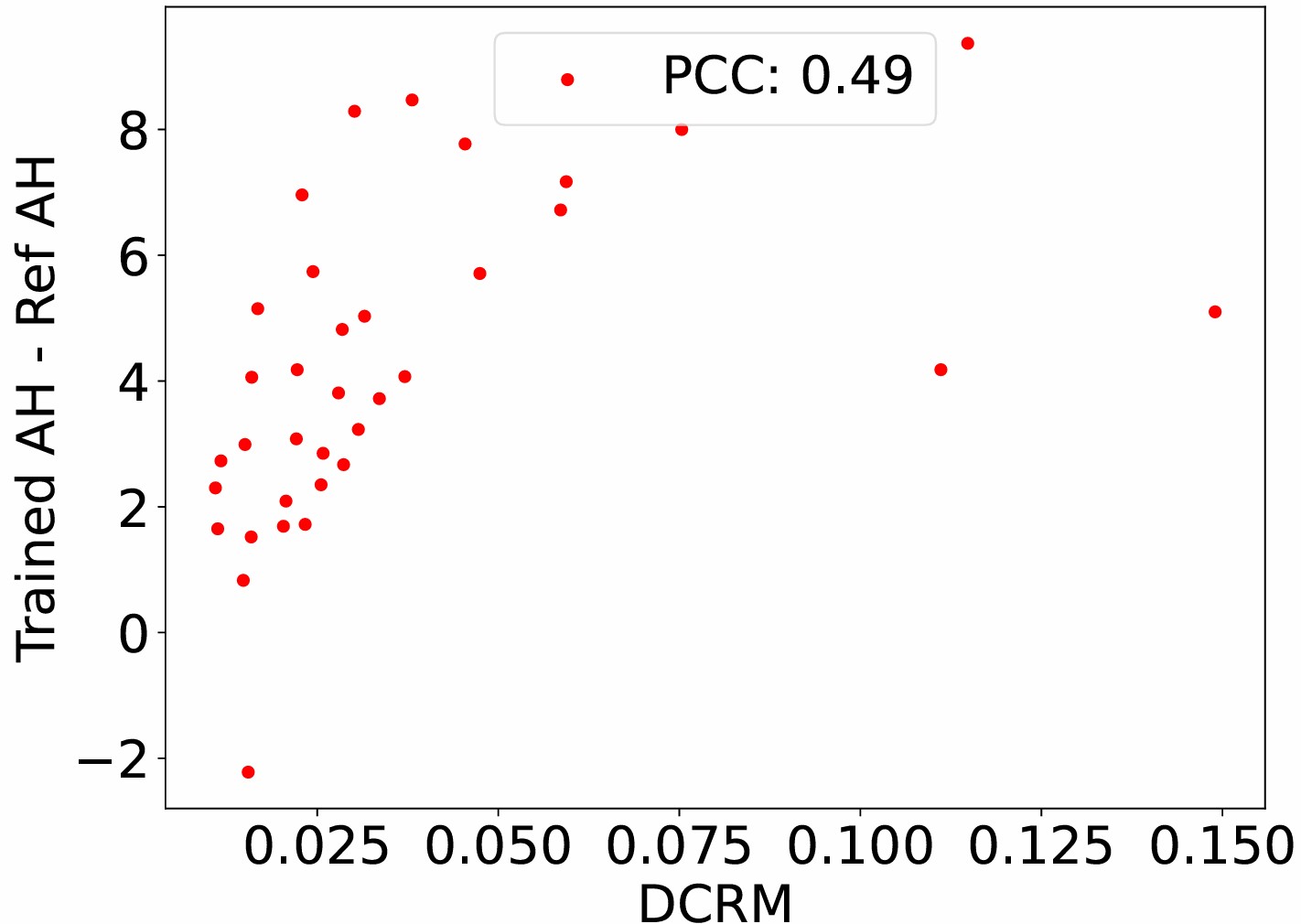}
    \caption{Correlation with Arena-Hard Performance}
    \label{fig:corr_ah}
\end{figure}

In Figure~\ref{fig:corr_mt} and \ref{fig:corr_ah} we show the correlations between \texttt{DCRM} and the model's performance changes on MT-Bench and Arena-Hard. We observe a weak positive correlation with MT-Bench scores with a Pearson Correlation Coefficient of 0.20, possibly due to the fact that MT-Bench evaluates the model's multi-turn conversation abilities, while our dataset and training are for single-turn conversation. Arena-Hard shows a moderate positive correlation, with a Pearson Correlation Coefficient of 0.49, similar to the case with AlpacaEval discussed in \S~\ref{sec:result}.

\section{Task-specific and OOD Downstream Performance}
\label{app:downstream_performance}

\begin{table}
  \small
  \centering
  
  \begin{tabular}{llcc}
    \toprule
    & & GSM8K & ME-Hard \\
    \midrule
    & \lmatwos & 23.28 & 24.55 \\
    \midrule
    SS-RM & +$\pi_{\mathrm{ref}}$ & 21.51 & 24.3 \\
    & w/ $BoN^2$ & 23.35 & 25.75 \\
    & +Mst & 21.76 & \textbf{26.5} \\
    & w/ $BoN^2$ & 22.14 & 25.4 \\
    & +Lma3 & \textbf{23.91} & 26.1 \\
    & +Gma2 & 23.76 & 26.1 \\
    \midrule
    DS-RM & +Gma2-Mst & 22.16 & 25.65 \\
    & w/ $BoN^2$ & 22.90 & \textbf{26.5} \\
    \midrule
    DS-Fix & +Gma2-Lma3 & 19.82 & 21.1 \\
    & +Gma2-Mst & 18.62 & 20.55 \\
    \bottomrule
  \end{tabular}
  \caption{\label{tab:downstream_performance}
    Task-specific and out-of-distribution downstream performance of each setting. GSM8K: 5-shot accuracy on GSM8K; ME-Hard: MixEval-Hard overall score; Training on DS-Fix datasets hurts models' performance while training on other datasets generally preserves or even increases the performance. 
  }
\end{table}

% GSM8K, MixEval:
% LLAMA-2-7B-Chat 7+3

We also investigate more task-specific and out-of-distribution downstream performance for each setting, using GSM8K \citep{cobbe2021}\footnote{We use the lm-evaluation-harness library version 0.4.5 at \url{https://github.com/EleutherAI/lm-evaluation-harness/tree/v0.4.5} to compute GSM8K results.} and MixEval-Hard \citep{ni2024}. As shown in Table~\ref{tab:downstream_performance}, the model's performance trained in the DS-Fix settings decreases compared with the base model $\pi_{\mathrm{ref}}$, while in other settings the performance is maintained close to $\pi_{\mathrm{ref}}$ or even increases. This suggests that noisy signals learned from the DS-Fix datasets not only hurt LLMs' general conversational abilities but also their task-specific downstream effectiveness.

\section{Ablation Study for best-of-$N^2$ pairing}
\label{app:ablation_study}

We also do an ablation study in the \lmatwos~$\pi_{\mathrm{ref}}$ (SS-RM) setting. In particular, we remove 1 of $e_\Delta$, $p_\Delta$, and $r_\Delta$ from DCRM. Removing $e_\Delta$ or $p_\Delta$ means setting DCRM's denominator to $p_\Delta+\epsilon$ or $e_\Delta+\epsilon$. Removing $r_\Delta$ means setting \texttt{DCRM} to just $\frac{1}{e_\Delta+p_\Delta+\epsilon}$, in which case the new Best of $N^2$ method effectively selects the pair with the smallest distance.

% AlpacaEval:
% LLAMA-2-7B-Chat SFT setting: ted, apd, ted+apd, ted+rw margin, apd+rw margin, ted+apd+rw margin

\begin{table}
  \small
  \centering
  
  \begin{tabular}{lccc}
    \toprule
     & AP-L & AP-R & Length\\
    \midrule
    \lmatwos & 12.57 & 10.43 & 1502\\
    \midrule
    +(SS-RM) $\pi_{\mathrm{ref}}$ & 22.36 & 16.81 & 1530\\
    w/ $BoN^2$ & 22.41 & 17.20 & 1561\\
    ~~~~-$p_\Delta$  & 22.1 & \textbf{17.27} & 1526\\
    ~~~~-$e_\Delta$  & \textbf{24.04} & 17.14 & 1513\\
    ~~~~-$r_\Delta$ & 14.81 & 12.11 & 1529 \\
    \bottomrule
  \end{tabular}
  \caption{\label{tab:ablation_1}
    Ablation Study on DCRM
  }
\end{table}

Table~\ref{tab:ablation_1} shows that \textbf{the performance after removing either $e_\Delta$ or $p_\Delta$ is close to that of the complete metric}. In Table~\ref{tab:ablation_2}, ~\ref{tab:ablation_3}, and ~\ref{tab:ablation_4} we have similar observations in other settings too. A merit entailed by this insight is that, in certain settings such as Mst (SS-RM) and DS-RM, our method can work well with just $e_\Delta$ and $r_\Delta$, without the need for a forward pass on the model to compute $p_\Delta$. $r_\Delta$ are usually collected during the preference annotation process and given in the preference dataset. In this case, we only need to compute $e_\Delta$ to apply our selection strategy, which is cheap and simple. 

\begin{table}[h]
  \small
  \centering
  % \resizebox{0.5\textwidth}{!}{
  \begin{tabular}{lccc}
    \toprule
     & AP-L & AP-R & Length\\
    \midrule
    \lmatwos & 12.57 & 10.43 & 1502\\
    \midrule
    +(SS-RM) $\pi_{\mathrm{ref}}$ & 22.36 & 16.81 & 1530\\
    w/ $BoN^2$ & 22.41 & 17.20 & 1561\\
    ~~~~-$p_\Delta$  & 22.1 & \textbf{17.27} & 1526\\
    ~~~~-$e_\Delta$  & \textbf{24.04} & 17.14 & 1513\\
    \midrule
    +(SS-RM) Mst & 15.49 & 12.07 & 1463\\
    w/ $BoN^2$ & \textbf{17.42} & \textbf{13.29} & 1456\\
    ~~~~-$p_\Delta$  & 16.86 & 12.80 & 1446\\
    ~~~~-$e_\Delta$  & 17.13 & 13.04 & 1446\\
    \midrule
    +(DS-RM) Gma2-Mst & 14.13 & 11.51 & 1511\\
    w/ $BoN^2$ & 16.82 & 13.6  & 1522\\
    ~~~~-$p_\Delta$  & 16.8 & 13.54 & 1528\\
    ~~~~-$e_\Delta$  & \textbf{17.54} & \textbf{13.98} & 1518\\
    \bottomrule
  \end{tabular}
  \caption{\label{tab:ablation_2}
    On \lmatwos, keeping $r_\Delta$ and 1 distance metric also works reasonably well and gives performance close to the complete metric.
  }
\end{table}

\begin{table}[h]
  \small
  \centering
  % \resizebox{0.5\textwidth}{!}{
  \begin{tabular}{lccc}
    \toprule
     & AP-L & AP-R & Length\\
    \midrule
    \gem & 16.07 & 10.31 & 1224\\
    \midrule
    +(SS-RM) $\pi_{\mathrm{ref}}$ & 27.03 & \textbf{18.01} & 1357\\
    w/ $BoN^2$ & \textbf{28.08} & 17.64 & 1343\\
    ~~~~-$p_\Delta$  & 26.73 & 16.02 & 1311\\
    ~~~~-$e_\Delta$  & 28.2 & 17.76 & 1331\\
    \midrule
    +(SS-RM) Mst & 22.96 & 14.66 & 1349\\
    w/ $BoN^2$ & \textbf{26.71} & \textbf{16.89} & 1328\\
    ~~~~-$p_\Delta$  & 25.37 & 16.67 & 1355\\
    ~~~~-$e_\Delta$  & 25.89 & 15.65 & 1278\\
    \midrule
    +(DS-RM) Lma3-Mst & 22.36 & 15.03 & 1379\\
    w/ $BoN^2$ & \textbf{26.14} & 17.76  & 1432\\
    ~~~~-$p_\Delta$  & 25.89 & \textbf{18.63} & 1458\\
    ~~~~-$e_\Delta$  & 24.12 & 15.78 & 1364\\
    \bottomrule
  \end{tabular}
  \caption{\label{tab:ablation_3}
    On \gemtwo, keeping $r_\Delta$ and 1 distance metric also works reasonably well and gives performance close to the complete metric.
  }
\end{table}

\begin{table}[h]
  \small
  \centering
  % \resizebox{0.5\textwidth}{!}{
  \begin{tabular}{lccc}
    \toprule
     & AP-L & AP-R & Length\\
    \midrule
    \lmathro & 14.15 & 15.34 & 1980\\
    \midrule
    +(SS-RM) $\pi_{\mathrm{ref}}$ & 22.8 & 25.65 & 2725\\
    w/ $BoN^2$ & \textbf{24.77} & 27.64 & 2825\\
    ~~~~-$p_\Delta$  & 24.58 & \textbf{27.89} & 2582\\
    ~~~~-$e_\Delta$  & 24.19 & 27.33 & 2716\\
    \midrule
    +(SS-RM) Mst & 19.43 & 19.94 & 1980\\
    w/ $BoN^2$ & \textbf{21.73} & \textbf{21.37} & 1915\\
    ~~~~-$p_\Delta$  & 21.08 & 20.56 & 1892\\
    ~~~~-$e_\Delta$  & 21.00 & 20.68 & 1882\\
    \midrule
    +(DS-RM) Gma2-Mst & 20.01 & 21.61 & 2062\\
    w/ $BoN^2$ & \textbf{24.53} & \textbf{27.76}  & 2145\\
    ~~~~-$p_\Delta$  & 23.43 & 26.52 & 2181\\
    ~~~~-$e_\Delta$  & 23.16 & 26.21 & 2127\\
    \bottomrule
  \end{tabular}
  \caption{\label{tab:ablation_4}
    On \lmathro, keeping $r_\Delta$ and 1 distance metric also works reasonably well and gives performance close to the complete metric.
  }
\end{table}

\begin{table}[h]
  \small
  \centering

  \begin{tabular}{lccc}
    \toprule
     & AP-L & AP-R & Length\\
    \midrule
    \lmatwos & 12.57 & 10.43 & 1502\\
    \midrule
    +~~(SS-RM) $\pi_{\mathrm{ref}}$ & 22.36 & 16.81 & 1530\\
    w/ $BoN^2$ & \textbf{22.41} & \textbf{17.20} & 1561\\
    ~~~~$e_\Delta$  only & 13.97 & 11.68 & 1538 \\
    ~~~~$p_\Delta$ only & 15.89 & 13.11 & 1537 \\
    ~~~~$e_\Delta$ +$p_\Delta$(DCRM-$r_\Delta$) & 14.81 & 12.11 & 1529 \\
    \bottomrule
  \end{tabular}
  \caption{\label{tab:ablation_5}
    Ablation Study on \texttt{DCRM} without reward margins. Selecting response pairs with the smallest distances leads to suboptimal performance.
  }
\end{table}

\paragraph{Removing $r_\Delta$ makes training less effective.} In general, we observe in Table~\ref{tab:ablation_5} that purely optimizing against distances with either $e_\Delta$, $p_\Delta$, or both is much less effective than when $r_\Delta$ is included. This is expected, since selecting the pair with the smallest distance reduces the reward margin significantly, indicating that not only the noisy differences but also the desired differences are eliminated in the selected pair.

\section{Discussion on Computational cost}
\label{app:computation_cost}
The term $N^2$ in $BoN^2$ comes from pairing the $N$ responses. We analyze the cost $BoN^2$ incurs and compare that with the cost of the conventional response pairing methods (e.g., selecting the response pair with the largest $r_\Delta$).

Firstly, in the sampling stage, similar to conventional methods, we sample $N$ responses (not $N^2$ responses) from the model, which means we do not incur extra sampling cost.

Secondly, in the reward scoring stage, we again follow conventional methods and use the reward model to give each response a reward score.

Thirdly, during pairing, we compute $r_\Delta$, $e_\Delta$, and $p_\Delta$ for each response pair. Computing $r_\Delta$ just needs simple arithmetic. For $p_\Delta$, again only simple arithmetic is needed, and the log probability of each response is a readily available byproduct when sampling the responses. $e_\Delta$ can be efficiently computed using existing libraries such as the edit distance library from Python. After this, we compute the \texttt{DCRM} value of each response pair and select the pair with the highest \texttt{DCRM} value.

Although the third stage incurs quadratic costs in terms of the number of responses $N$, we argue that these costs are still minimal, since (1) a relatively small $N$ is usually sufficient and large $N$ gives diminishing returns (See Appendix~\ref{app:increasing_n}), (2) the cost for each pair is minimal either due to arithmetic simplicity ($r_\Delta$ and $p_\Delta$) or implementation efficiency ($e_\Delta$).

Therefore, the bulk of computation is still spent on response sampling and reward scoring, which are the same in our and conventional methods, and applied to each output separately (i.e., O(N) compute). Consequently, the total extra cost our method incurs is comparable to the conventional methods.

\section{Increasing Number of Responses}
\label{app:increasing_n}
The hyperparameter $N$ controls the number of responses in the response pool. Increasing $N$ should help create a more diverse set of responses, boost the quality of the response pairs identified by our $BoN^2$ method, and raise the trained model's performance on benchmarks. To inspect the effect of increasing $N$, we change $N$ from the original value of 5 to 8 and curate a new dataset to train \lmatwos. Table~\ref{tab:increasing_n} shows the results.

\begin{table}[h]
  \small
  \centering

  \begin{tabular}{lccc}
    \toprule
     & AP-L & AP-R & Length\\
    \midrule
    \lmatwos & 12.57 & 10.43 & 1502\\
    \midrule
    +~~(SS-RM) $\pi_{\mathrm{ref}}$ & 22.36 & 16.81 & 1530\\
    w/ $BoN^2, N=5$ & 22.41 & 17.20 & 1561\\
    w/ $BoN^2, N=8$ & \textbf{23.35} & \textbf{17.89} & 1548\\
    \bottomrule
  \end{tabular}
  \caption{\label{tab:increasing_n}
    Results with different values of $N$. Increasing $N$ beyond 5 gives diminishing returns.
  }
\end{table}

As $N$ increases, we observe diminishing returns. We suspect the reason to be that we are only sampling from one single source model, which puts an upper bound on the response diversity. However, this is not a deficiency specific to our method. In fact, all methods that sample multiple responses from the same model will eventually suffer from diminishing returns as $N$ increases.

% \section{Example Appendix}
% \label{sec:appendix}

% This is an appendix.

\end{document}